\documentclass[journal]{IEEEtran}
\usepackage{amssymb,amsfonts,amsmath,amscd}
\usepackage[ruled,vlined,linesnumbered]{algorithm2e}  %
\usepackage{array}
\usepackage[caption=false,font=footnotesize,labelfont=rm,textfont=rm]{subfig}
\usepackage{textcomp}
\usepackage{stfloats}
\usepackage{url}
\usepackage{verbatim}
\usepackage{graphicx}
\usepackage{subfig}
\usepackage{cite}
\usepackage{hyperref}
\hypersetup{
colorlinks=true,
allcolors=blue,
linkcolor=blue,
anchorcolor=blue,
citecolor=blue}
\usepackage{threeparttable}

\newcommand{\setInsert}[0]{\xleftarrow{\scriptscriptstyle +}}
\newcommand{\setRemove}[0]{\xleftarrow{\scriptscriptstyle -}}

\newcommand{\algo}[1]{Alg.~\ref{#1}}
\newcommand{\algoline}[2]{\algo{#1}, Line~\ref{#2}}
\newcommand{\algolines}[3]{\algo{#1}, Lines~\ref{#2}--\ref{#3}}

\DeclareMathAlphabet{\mathscr}{OT1}{pzc}{m}{it}

\begin{document}
%
\title{PPNet: A Two-Stage Neural Network \\for End-to-end Path Planning}
%
%

\author{Qinglong~Meng,
        Chongkun~Xia,
        Xueqian~Wang,
        Songping~Mai,
        and~Bin~Liang,
        \thanks{
        This work was supported by the National Key R\&D Program of China (2022YFB4701400/4701402), National Natural Science Foundation of China (No. U21B6002, 62203260, 92248304), Guangdong Basic and Applied Basic Research Foundation (2023A1515011773), Science, Technology and Innovation Commission of Shenzhen Municipality (WDZC20200820160650001 \& JCYJ20200109143003935).
        \par Qinglong Meng, Chongkun Xia, Xueqian Wang, and Songping Mai are with Tsinghua Shenzhen International Graduate School, Shenzhen 518055, China (e-mail: mengql22@mails.tsinghua.edu.cn; xiachongkun@qq.com; wang.xq@sz.tsinghua.edu.cn; mai.songping@sz.tsinghua.edu.cn)).
        \par Bin Liang is with the Department of Automation, Tsinghua University, Beijing 100854, China (e-mail: bliang@mail.tsinghua.edu.cn).}}

%
%

\markboth{Journal of \LaTeX\ Class Files,~Vol.~14, No.~8, August~2015}%
{Shell \MakeLowercase{\textit{et al.}}: Bare Demo of IEEEtran.cls for IEEE Journals}
%



\maketitle

\begin{abstract}
The classical path planners, such as sampling-based path planners, can provide probabilistic completeness guarantees in the sense that the probability that the planner fails to return a solution if one exists, decays to zero as the number of samples approaches infinity. However, finding a near-optimal feasible solution in a given period is challenging in many applications such as the autonomous vehicle. To achieve an end-to-end near-optimal path planner, we first divide the path planning problem into two subproblems, which are path space segmentation and waypoints generation in the given path's space. We further propose a two-stage neural network named Path Planning Network (PPNet) each stage solves one of the subproblems abovementioned. Moreover, we propose a novel efficient data generation method for path planning named EDaGe-PP. EDaGe-PP can generate data with continuous-curvature paths with analytical expression while satisfying the clearance requirement. The results show the total computation time of generating random 2D path planning data is less than 1/33 and the success rate of PPNet trained by the dataset that is generated by EDaGe-PP is about $2 \times$ compared to other methods. We validate PPNet against state-of-the-art path planning methods. The results show that PPNet can find a near-optimal solution in 15.3ms, which is much shorter than the state-of-the-art path planners.
\end{abstract}

\def\abstractname{Note to Practitioners}
\begin{abstract}
This article aims to provide an end-to-end near-optimal path planning method for applications such as autonomous driving, warehouse robot, and others. Sampling-based methods are the popular algorithms in these areas due to their good scalability and high efficiency. But the quality of the path that these methods find in a relatively short planning time can not be guaranteed. To guarantee the quality of the path which is found in a short period of time, we propose a neural network named PPNet. It can find a path with quality guarantee by one time of forward propagation. 
\end{abstract}


%
\IEEEpeerreviewmaketitle

\section{Introduction}\label{sec:intro}
\IEEEPARstart{P}{ath} planning is one of the core research problems in robotics. It is to find a collision-free, low-cost path connecting the initial state and goal state. The popular methods, such as sampling-based path planners, can provide probabilistic completeness guarantees, which means the probability that the planners succeed in returning a solution increases by increasing the number of samples~\cite{rrt*}. Therefore, they can't guarantee to find a near-optimal solution within a specific time, especially when the required planning time is quite short (i.e., $\leq$0.1s) in many applications such as the autonomous vehicle. Thus many variants of RRT* were proposed to shorten computation time. For example, Informed RRT* (IRRT*)~\cite{irrt*} is a representative variant of implementing Informed Search on RRT*. After finding the initial solution, IRRT* can converge to the optimal solution more quickly. Batch Informed Trees (BIT*)~\cite{bit*}, and Advance BIT* (ABIT*)~\cite{abit*} are to shorten computation time by further using the heuristics function to direct the process of exploring state space. It makes the methods find the initial solution more quickly in some cases. However, the heuristic sampling domain of the abovementioned methods might be quite large in some cases, and sampling in the domain remains random in these methods. The inefficiency of exploring the state space makes the computation time of the methods might be unacceptable for the applications mentioned above.
\par Recently, machine learning techniques have been used in path planning to overcome the limitations of sampling-based path planners. Many learning-based methods can perform quite short computation time in the learned environments. Neural RRT* (NRRT*)~\cite{nrrt*} combines the machine learning techniques and the sampling-based method. NRRT* uses a neural network for segmenting the promising region in the state space. Letting RRT* sample in the segmented region, which is smaller than the sampling domain of the methods used informed search, can accelerate the process of finding the initial path and converging to the optimal path. Motion Planning Networks (MPNet)~\cite{mpnet} uses RNN for predicting the next waypoint directly. Without random sampling, MPNet can lay out a near-optimal path with the necessary iterations in some cases. However, these works still need to increase iterations to improve the success rate, which makes computation time uncontrollable. 
\par
\begin{figure*}[t!]
	\centering
	\subfloat[\label{fig.optimal.ppnet}]{
		\includegraphics[height=0.24\textwidth]{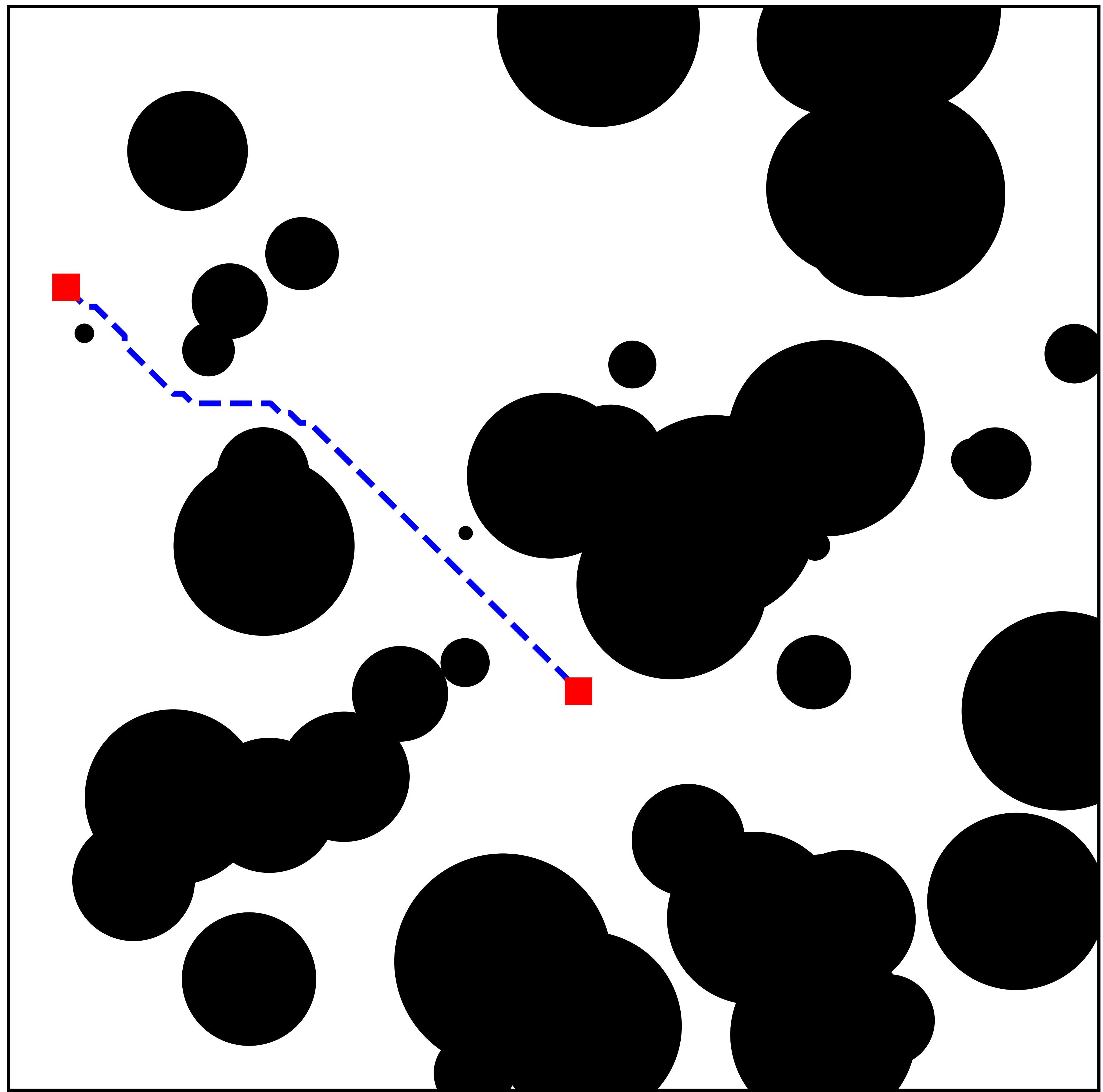}}
	\subfloat[\label{fig.optimal.rrt15}]{
		\includegraphics[height=0.24\textwidth]{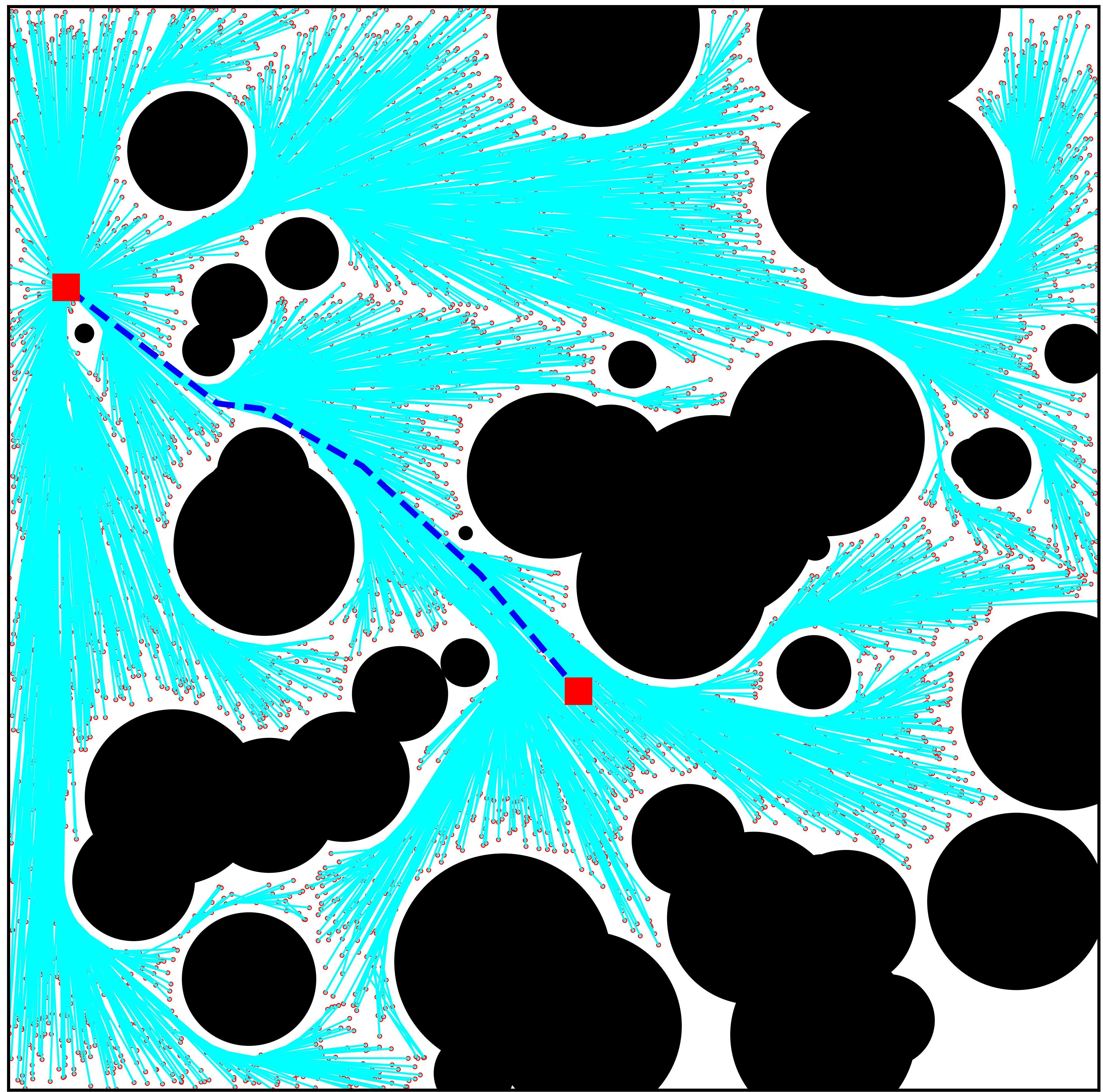}}
	\subfloat[\label{fig.optimal.rrt30}]{
		\includegraphics[height=0.24\textwidth]{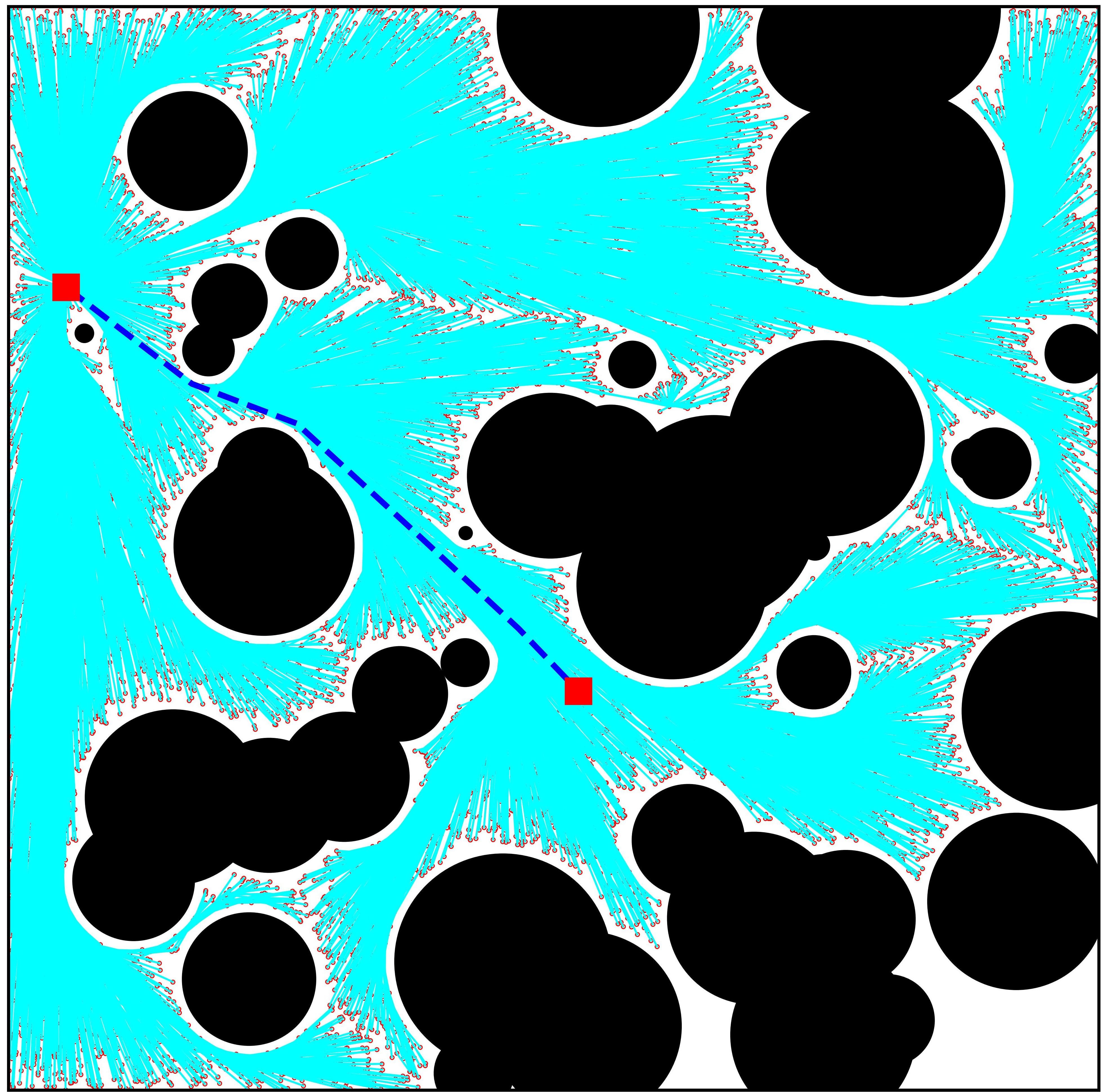}}
  	\subfloat[\label{fig.optimal.rrt60}]{
		\includegraphics[height=0.24\textwidth]{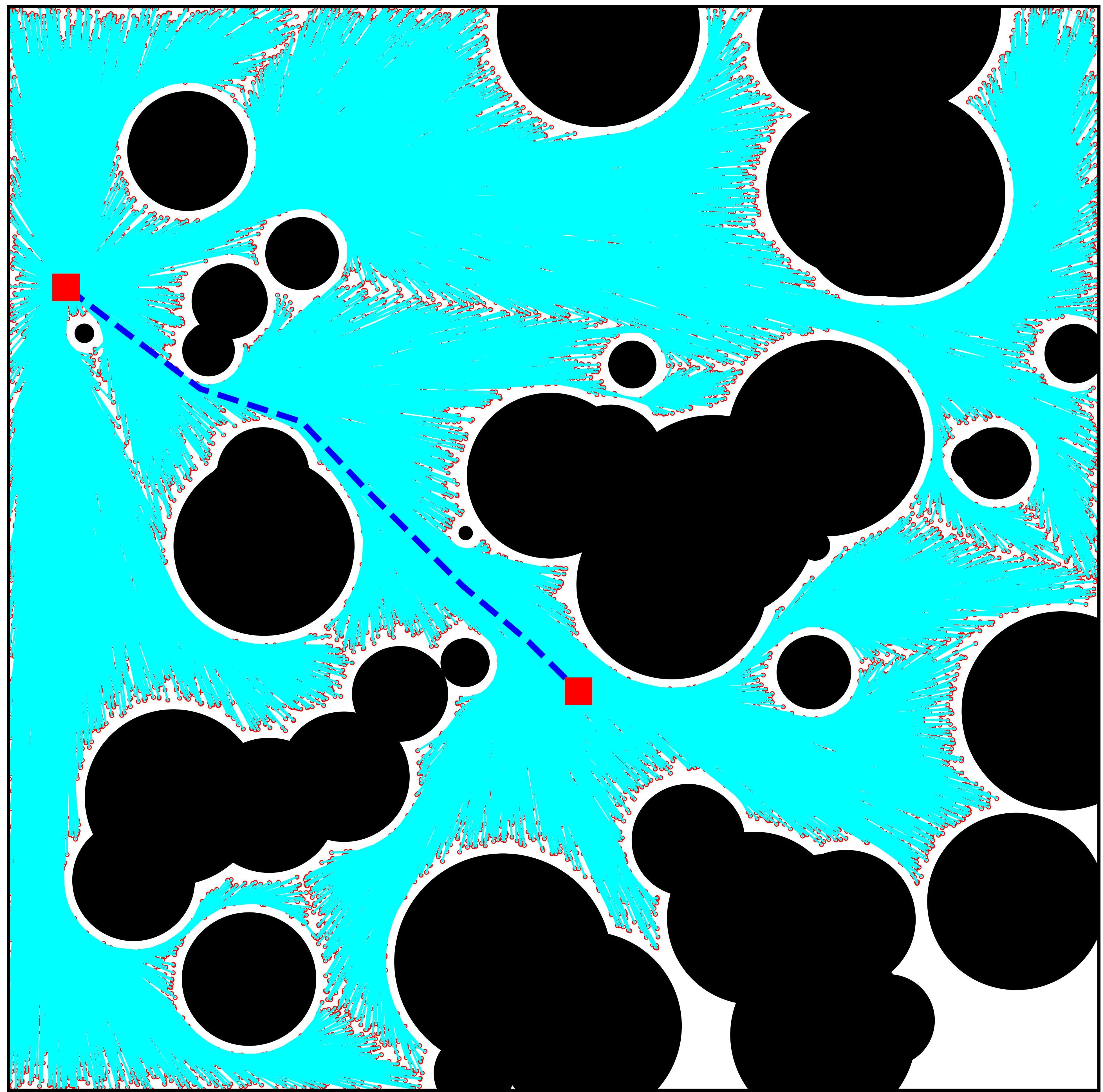}}
    \caption{PPNet can find a near-optimal solution in an end-to-end way. The representative classical planner\textemdash RRT* can find a similarly optimal solution of Euclidean cost within the 5\% range of the cost of the PPNet solution after the tree is expanded to have over 20,000 vertices. (a) PPNet, $t=0.015 \mathrm{s}$, $c=31.49$. (b) RRT*, $t=15.002 \mathrm{s}$, $c=30.69$. (c) RRT*, $t=30.001 \mathrm{s}$, $c=30.41$. (d) RRT*, $t=60.004 \mathrm{s}$, $c=30.48$. }
    \label{fig.optimal}
\end{figure*}

 To overcome the limitation abovementioned, we think that developing an end-to-end near-optimal path planner is a promising way that provide theoretical guarantees for finding a near-optimal path in a short period. To achieve a method that can find a near-optimal solution in an end-to-end way, we first divide the path planning problem into two subproblems, which are path space segmentation and waypoints generation in the given path's space. We further propose a novel two-stage neural network named Path Planning Network (PPNet) that can solve the path planning problem by solving the two subproblems orderly. PPNet comprises a segmentation model as the first stage named SpaceSegNet and a generation model as the second stage named WaypointGenNet. SpaceSegNet takes the map with the initial point and goal point mark as input and outputs the path space. WaypointGenNet takes the output of SpaceSegNet as input and outputs the probability map indicating the probability of whether each point in the environment is the waypoint. With the probability map, the path can be extracted based on a simple rule during the computation time of PPNet. Moreover, to achieve a better success rate of PPNet, we also propose a novel efficient data generation method for path planning named EDaGe-PP. The results show that EDaGe-PP provides about $33\times$ computation speed improvement and $2\times$ success rate improvement compared with the popular methods. The data generated by EDaGe-PP and the two-stage structure of PPNet enable PPNet to achieve about $41\%$ success rate improvement compared with the representative learning-based method and find a near-optimal path in 15.3ms (Fig.~\ref{fig.optimal}).
\par The remaining article is organized as follows. Section~\ref{sec:bg} reviews the main pieces of classical and deep-learning-based methods. Section~\ref{sec:data} presents the details of EDaGe-PP. Section~\ref{sec:net} presents the details of PPNet. Section~\ref{sec:exp} presents experimental results in detail. Section~\ref{sec:clusion} concludes the paper with a discussion of the technique.

\section{Related Work}\label{sec:bg}
 The performance requirement that some applications such as the autonomous vehicle require a near-optimal path in a specific time is challenging to the computational efficiency of path planners. However, due to the randomly ordered search, popular methods such as sampling-based methods (e.g., RRT* and PRM*) might not satisfy such a performance requirement. Two reasons for the inefficiency of randomly ordered search are random sampling in the whole state space and expansion of the tree that is guided by the random samples. To improve the efficiency of expanding the tree, Fast Marching Tree (FMT*)~\cite{fmt*} uses a marching method to process samples. But for achieving a 100\% success rate of FMT*, the number of samples should be larger than 1000 and the execution time might be longer than 1s~\cite{fmt*}. If the number of samples is less than that, the success rate will decrease with the number of samples rapidly. The methods used Informed search~\cite{irrt*} sample in the heuristic sampling domain, which is determined by a hypothesis or information that has been known. Instead of sampling in the whole state space, the methods improve the convergence to the optimal path. But the sampling domain might be quite large in some cases, and the expansion of the tree is still inefficient. BIT*~\cite{bit*} further uses incremental graph-search techniques like Lifelong Planning A* (LPA*)~\cite{lpa*} to guide the expansion of the tree. ABIT*~\cite{abit*} achieves a shorter time of finding an initial path by using more advanced graph-search techniques, such as inflation and truncation. It means BIT* and ABIT* are to shorten the computation time by improving both the sampling and the expansion. However, these improvements might be insufficient for some applications because the heuristic function that directs the expanding of the tree makes these methods might get trapped in some cases.
\par By applying the machine learning techniques there are methods that achieve ordered search. NEED~\cite{need} uses a segmentation model to segment the promising region of the state space as the preprocessing for sampling-based methods. The promising region is the region that has a high probability that waypoints are in it. Neural RRT*~\cite{nrrt*} and 3D Neural RRT*~\cite{3dnrrt*} can be seen as the application of NEED which combines RRT* with NEED. Neural RRT* and 3D Neural RRT* improve the efficiency of search by searching in the promising region which can be seen as the variant of informed search. But instead of searching in the ellipse that is drawn by the cost of the initial solution, they search in the promising region which is segmented by the segmentation model. The more narrow sampling region makes the sampling more efficient. MPNet~\cite{mpnet} avoids random sampling and expanding by using a recurrent neural network (RNN) to output the next waypoint directly. And due to bidirectional planning like RRT-Connect~\cite{rrt-c}, MPNet alleviates the difficulty of predicting the long sequence of RNN. MPNet could lay out the near-optimal solution in some learned environments without generating extra nodes. But in complex environments the necessary number of iterations of the abovementioned methods might be quite large, and for improving the quality of the path the number needs to be further increased.
\par Unlike the previous works that apply machine learning techniques, PPNet solves the path planning problem by solving the two subproblems orderly, which makes PPNet able to output a path in an end-to-end way. Moreover, the continuous-curvature paths with analytical expression, which is generated by EDaGe-PP, make PPNet achieve more efficient training and a much higher success rate.

\section{EDaGe-PP: Efficient Data Generation}\label{sec:data}
In this section, we describe EDaGe-PP. EDaGe-PP was proposed for generating the custom data more efficiently for the training of learning-based path planning methods. It can overcome four limitations of the popular data generation methods for learning-based methods. Firstly, using classical planners to generate an optimal path is extremely time-consuming in complex environments. Secondly, the paths only contain a few waypoints is computationally expensive to generate data with the custom formats for the training of neural networks, which is usual in developing learning-based methods. Thirdly, they both need appropriate \emph{a prior} discretization for the path planning problem with clearance requirements, which is difficult sometimes. Fourthly, the paths comprised of discrete waypoints make it difficult for the neural networks to converge. To overcome the limitations of popular data generation methods mentioned above, we propose an efficient data generation method for path planning named EDaGe-PP. EDaGe-PP can generate data with continuous-curvature paths with analytical expression while satisfying the clearance requirement. Moreover, EDaGe-PP provides about 33× computation speed improvement and 2× success rate improvement compared with the popular methods. 
\par In this paper, we use a 2D RGB image to represent the state space, which is denoted as $\mathscr{R}$, and use the white and black colors respectively to indicate the free space and the obstacle space. The initial state and goal state are both represented by red rectangles. The data generated by EDaGe-PP is comprised of an image representing the path planning problem, a mask representing the space of the path with clearance requirement, and a mask representing the waypoints of the path. EDaGe-PP is presented in \algo{alg:data}. The data is generated by 4 steps in our algorithm. The first step is to randomly generate paths that will be the near-optimal solutions in the state spaces, and the original masks representing the waypoints of the paths (\algoline{alg:data}{alg:data:3}), which is described in Section~\ref{sec:data:path}. The second step is to calculate the space of the path with the clearance requirement (\algoline{alg:data}{alg:data:5}), which is described in \algo{alg:space}. The third step is to calculate the settings of obstacles that make the path a near-optimal solution in the state space with these settings (\algoline{alg:data}{alg:data:6}), which is described in \algo{alg:obs}. The last step is to add the settings which are calculated in the second step and the random obstacles are collision-free with the path (\algoline{alg:data}{alg:data:9}), which is described in \algo{alg:prob}.

\begin{algorithm}[t]
    \caption{EDaGe-PP$\left ( c, n_{P}, n_{I}\right )$}
    \label{alg:data}
    $\mathcal{P}\gets \emptyset;\mathcal{D}\gets \emptyset;O\gets \emptyset;$\label{alg:data:1}\\
    $\mathcal{I}_{P}\gets \emptyset;\mathcal{I}_{Path}\gets \emptyset;\mathcal{I}_{Space}\gets \emptyset;\mathcal{I}_{Problem}\gets \emptyset;$\label{alg:data:2}\\
    $\mathcal{P}, \mathcal{I}_{P} = \mathrm{GeneratePath}(n_{P})$\label{alg:data:3}\\
    \ForEach{$P$, $I_{P}$ in $\mathcal{P}$, $\mathcal{I}_{P}$}{\label{alg:data:4}
        $B, I_{S} = \mathrm{CalcuSpace(c, P)}$\label{alg:data:5}\\
        $O = \mathrm{SetObstacles(c, P)}$\label{alg:data:6}\\
        $t=0$\label{alg:data:7}\\
        \Repeat{$t \le n_{I}$}{\label{alg:data:8}
            $\mathcal{I}_{Path}, \mathcal{I}_{Space},\mathcal{I}_{Problem} \setInsert \mathrm{RandomStateSpace}(B,O,I_{S},I_{P})$\label{alg:data:9}
        }
    }
    \Return{$\mathcal{I}_{Path}, \mathcal{I}_{Space},\mathcal{I}_{Problem}$}\;\label{alg:data:11}
\end{algorithm}

\begin{algorithm}[t]
    \caption{CalcuSpace$\left ( c, P\right)$}
    \label{alg:space}
    $I_{S} \gets black$;\label{alg:space:1}\\
    $B=CalcuBoundary(P)$;\label{alg:space:2}\\
    \ForEach{$_{}^{u}\vec{b}$, $_{}^{l}\vec{b}$, $_{}^{i}\vec{b}$, $_{}^{e}\vec{b}$ in $_{}^{u}B$, $_{}^{l}B$, $_{}^{i}B$, $_{}^{e}B$}{\label{alg:space:3}
        $I_{S} = SetFree\left (I_{S}, _{}^{u}\vec{b}, _{}^{l}\vec{b} \right )$;\label{alg:space:4}\\
        $I_{S} = SetFree\left (I_{S}, \vec{i}, _{}^{i}\vec{b} \right )$;\label{alg:space:5}\\
        $I_{S} = SetFree\left (I_{S}, \vec{e}, _{}^{e}\vec{b} \right )$;\label{alg:space:6}\\
    }
    \Return{$B,I_{S}$}\;
\end{algorithm}

\begin{algorithm}[t]
    \caption{SetObstacles$\left ( c, P \right )$}
    \label{alg:obs}
    $O\gets \emptyset;$\\
    $H_{P}=ConvexHull(P);$\label{alg:obs:hull}\\
    $S=FindSubSpaces(H_{P});$\label{alg:obs:sub}\\
    \ForEach{$s$ in $S$}{\label{alg:obs:4}
        $\Vec{z}, \Vec{n}=CalcuDirection(s);$\label{alg:obs:5}\\
        $w=CalcuWidth(s);$\label{alg:obs:6}\\
        \Repeat{$\sum_{i=1}^{n} O[i].s \ge 2w$}{\label{alg:obs:7}
            \textbf{select randomly } 
            $\begin{cases}
            radius \in \left ( 0, 2w \right ]\\   
            \varepsilon_{n} \in \left ( 0, 1 \right ]\\  
            \varepsilon_{z} \in \left [ -1, 1 \right ]
            \end{cases};$\\\label{alg:obs:8}
            \If{$O == \emptyset$}{\label{alg:obs:9}
            $t_{n} = c$;\label{alg:obs:10}
            }
            \Else{
            $t_{n} = \varepsilon_{n} \cdot (radius + O[-1].radius)$;\label{alg:obs:12}
            }
            $t_{z} = \varepsilon_{z} \cdot radius$;\label{alg:obs:13}\\
            $\vec{t}= t_{n}\Vec{n}+ t_{z}\Vec{h}$; \label{alg:obs:14}\\
            $\vec{coord}=\vec{t}+O[-1].\vec{coord}$;\label{alg:obs:15}\\
            \If{$CollisonFree(P,\left [radius,\vec{coord}\right ])$}{\label{alg:obs:16}
                $O\setInsert \left [radius,\vec{coord}\right ]$;\label{alg:obs:17}
            }
        }
    }
    \Return{$O$}\;
\end{algorithm}

\begin{algorithm}[t]
    \caption{RandomStateSpaces$\left (B,O,I_{S},I_{P} \right )$.}
    \label{alg:prob}
    $H_{B}=ConvexHull(B);$\label{alg:prob:1}
    \Repeat{times exceeded}{\label{alg:prob:2}
        \textbf{select randomly } \label{alg:prob:3}
            $\begin{cases}
            \vec{t} \in \mathscr{R} \\
            r \in \left [ 0, 360^{\circ} \right ]
            \end{cases};$
        $R=\begin{bmatrix}\label{alg:prob:4}
                \cos r & -\sin r  \\
                \sin r & \cos r 
                \end{bmatrix}$\; 
        $H^{'}_{B}=R\cdot H_{B}+\vec{t}$\label{alg:prob:5}\;
        \If{$BoudaryCheck(H^{'}_{B})$}{\label{alg:prob:6}
            \textbf{select randomly } $O^{'} \in \mathscr{R}$\;\label{alg:prob:7}
            \ForEach{$o^{'}$ in $O^{'}$}{\label{alg:prob:8}
                \If{ not $CollisonFree(P,o^{'})$}{\label{alg:prob:9}
                    $O^{'}\setRemove o^{'}$\;\label{alg:prob:10}
                }
            }
            \ForEach{$o$ in $O$}{\label{alg:prob:11}
                $o.\vec{coord} = R\cdot o.\vec{coord} + \vec{t}$\;\label{alg:prob:12}
                $O^{'}\setInsert o$\;\label{alg:prob:13}
            }
            $I_{Prob} = GenerateMap(O^{'})$\;\label{alg:prob:14}
            $I_{S} = R\cdot I_{S} + \vec{t}$\;\label{alg:prob:15}
            $I_{P} = R\cdot I_{P} + \vec{t}$\;\label{alg:prob:16}
            $I_{Prob} = I_{Prob} \odot I_{S}$\;\label{alg:prob:17}
            break\;\label{alg:prob:18}
        }
    }
    \If{$Success$}{\label{alg:prob:20}
        \Return{$I_{P},I_{S},I_{Prob}$}\;\label{alg:prob:21}
    }
    \Else{\label{alg:prob:22}
        \Return{$Failure$}\;\label{alg:prob:23}
    }
\end{algorithm}

\subsection{Random Path Generation}\label{sec:data:path}
In this paper, we use a predetermined number of polynomial functions to describe the paths that are randomly generated by our algorithm. The polynomial function is defined as follows:
\begin{gather}
y=w^{\mathrm{T}} \cdot \vec{X},\\
\vec{w} =\left [ w_{0}, w_{1},\cdots  ,w_{r}  \right ]^{\mathrm{T}},\\
\vec{X} =\left [ 1,x,x^{2},\cdots  ,x^{r}  \right ]^{\mathrm{T}},
\end{gather}
where $r$ denotes the predetermined order of the polynomial function.
\par The coefficients of the polynomial function are randomly generated. But for the diversity of polynomial curves the coefficients are not sampled from a uniform distribution, the random coefficients are obtained from the fitting result for random samples from a uniform distribution.
\par Then waypoints are sampled from the polynomial curve at an equal distance. The matrix of the waypoints of each polynomial curve $P^{(i)}$ is defined as:
\begin{gather}
P^{(i)}=\left[\vec{p}_{1}^{(i)}, \vec{p}_{2}^{(i)}, \cdots, \vec{p}_{n}^{(i)}\right],\\
\vec{p}_{j}^{(i)}=\left[x_{j}^{(i)}, y_{j}^{(i)}\right]^{\mathrm{T}},
\end{gather}
where $i=1,2,\cdots,l$ denotes the index of the polynomial curve, $l$ denotes the number of polynomial curves, and $n$ denotes the number of the waypoints of each polynomial curve.
\par For concatenating the polynomial curves, the gradients of the first and last points of $l$ curves ($l$ denotes the predetermined number of polynomial curves) need to be calculated for the transformation. The rotation matrix $R^{(i)}$ and translation vector $\vec{t^{(i)}}$ are defined as:
\begin{gather}
R^{(i)}=\begin{bmatrix}
  \cos \alpha^{(i)}& -\sin \alpha^{(i)} \\
  \sin \alpha^{(i)} & \cos \alpha^{(i)} 
\end{bmatrix},\\
\alpha^{(i)}=\tan^{-1} \nabla y_{n}^{(i-1)} - \tan^{-1} \nabla y_{1}^{(i-1)},\\
\vec{t^{(i)}}=\vec{p}_{n}^{(i-1)}-\vec{p}_{1}^{(i)},
\end{gather}
\vspace{-15pt}
\\where $\alpha^{(i)}$ denotes the rotation angle of \emph{i}-th polynomial curve, $\sin$/$\cos$ denotes the sine/cosine function, and $\tan^{-1}$ denotes the arctangent function.
\par The waypoints of the path can be obtained by concatenating the $l$ polynomial curves. The waypoints of the path are defined as:
\begin{gather}
P=\left [ P_{T}^{(1)},P_{T}^{(2)},\cdots ,P_{T}^{(l)}  \right ],\\
P_{T}^{(i)}=R^{(i)}\cdot P^{(i)}+\vec{t}^{(i)},
\end{gather}
where $P$ denotes the matrix of the waypoints of the path, and $P_{T}^{(i)}$ denotes the transformed $P^{(i)}$.
\par After concatenating the polynomial curves, the path's waypoints can be used for generating the mask of the waypoints, which is represented by $I_{P}$.

\subsection{Space of the Path}\label{sec:data:space}
The algorithm for calculating the path's space is presented in \algo{alg:space}. 
With the polynomial functions of the path, the boundary of the path's space can be calculated. The boundary of the path's space is defined as (\algoline{alg:space}{alg:space:2}):
\begin{gather}
B =\left [ _{}^{u}B,_{}^{l}B,_{}^{i}B,_{}^{e}B \right ] ,
\end{gather}
where $_{}^{i/e}B$ denotes the boundary around initial/end point, and $_{}^{u/l}B$ denotes upper/lower boundary.
\par The upper/lower boundary of each polynomial curve can be calculated by the normal vector of the corresponding waypoint. The upper/lower boundary is defined as:
\begin{gather}
_{}^{u/l}B =\left [ _{}^{u/l}B_{T}^{(1)},_{}^{u/l}B_{T}^{(2)},\cdots,_{}^{u/l}B_{T}^{(l)}  \right ], \\
_{}^{u/l}B_{T}^{(i)}=R^{(i)} \cdot _{}^{u/l}B^{(i)}+\vec{t}^{(i)},\\
 _{}^{u/l}B^{(i)}=\left [ _{}^{u/l}\vec{b}_{1}^{(i)},_{}^{u/l}\vec{b}_{2}^{(i)},\cdots,_{}^{u/l}\vec{b}_{n}^{(i)} \right ] ,\\
_{}^{u/l}\vec{b}_{j}^{(i)}=\vec{p}_{j}^{(i)}\pm c \vec{n}_{j}^{(i)}, j=1,2,\cdots,n,\\
\vec{n}_{j}^{(i)} = \frac{\left [ \nabla y_{j}^{(i)},-1  \right ] }{\left \| \left [ \nabla y_{j}^{(i)},-1  \right ] \right \| } ,
\end{gather}
where $c$ denotes the width of the required clearance.
\par The boundary around the initial/end point is a circle that concatenates with the upper and lower boundary. The initial/end point is defined as:
\begin{gather}
_{}^{i/e}B =\left [ _{}^{i/e}\vec{b}_{1},_{}^{i/e}\vec{b}_{2},\cdots,_{}^{i/e}\vec{b}_{n^{'}} \right ] ,\\
_{}^{i/e}\vec{b}_{j}=\vec{i}/\vec{e}\pm c _{}^{i/e}R_{j}\cdot _{}^{i/e}\vec{t}, j=1,2,\cdots,n^{'},\\
_{}^{i/e}R_{j}=\begin{bmatrix}
  \cos _{}^{i/e}\beta _{j}& -\sin _{}^{i/e}\beta _{j} \\
  \sin _{}^{i/e}\beta _{j} & \cos _{}^{i/e}\beta _{j} 
\end{bmatrix},\\
_{}^{i/e}\beta _{j} = \frac{j_{}^{i/e}\vec{t}\cdot _{}^{i/e}\vec{t}^{'}}{n^{'}\left \| _{}^{i/e}\vec{t} \right \|\left \| _{}^{i/e}\vec{t}^{'} \right \|} ,\\
_{}^{i}\vec{t} = _{}^{u/l}\vec{b}_{1}^{(1)} - \vec{i},\\
_{}^{i}\vec{t}^{'} = _{}^{l/u}\vec{b}_{1}^{(1)} - \vec{i},\\
_{}^{e}\vec{t} = _{}^{u/l}\vec{b}_{n}^{(l)} - \vec{e},\\
_{}^{e}\vec{t}^{'} = _{}^{l/u}\vec{b}_{n}^{(l)} - \vec{e},
\end{gather}
where $n^{'}$ denotes the number of samples of the boundary around the initial point or the end point.
\par With the boundary of the path's space, we can consider $_{}^{u}\vec{b}_{j}^{(i)}$ and $_{}^{l}\vec{b}_{j}^{(i)}$, $_{}^{i}\vec{b}_{j}$ and $\vec{i}$, $_{}^{e}\vec{b}_{j}$ and $\vec{e}$ as pairs of points for setting free space of the state space. Between each pair of points, the space should be set to free for generating the image $I_{S}$ representing the path's space (\algolines{alg:space}{alg:space:4}{alg:space:6}). Since the path is the analytical solution, the mask of the path's space can be generated with clearance requirements.

\begin{figure*}[!t]
    \centering
	\subfloat[\label{fig.data.c1.1}]{
		\includegraphics[height=0.24\textwidth]{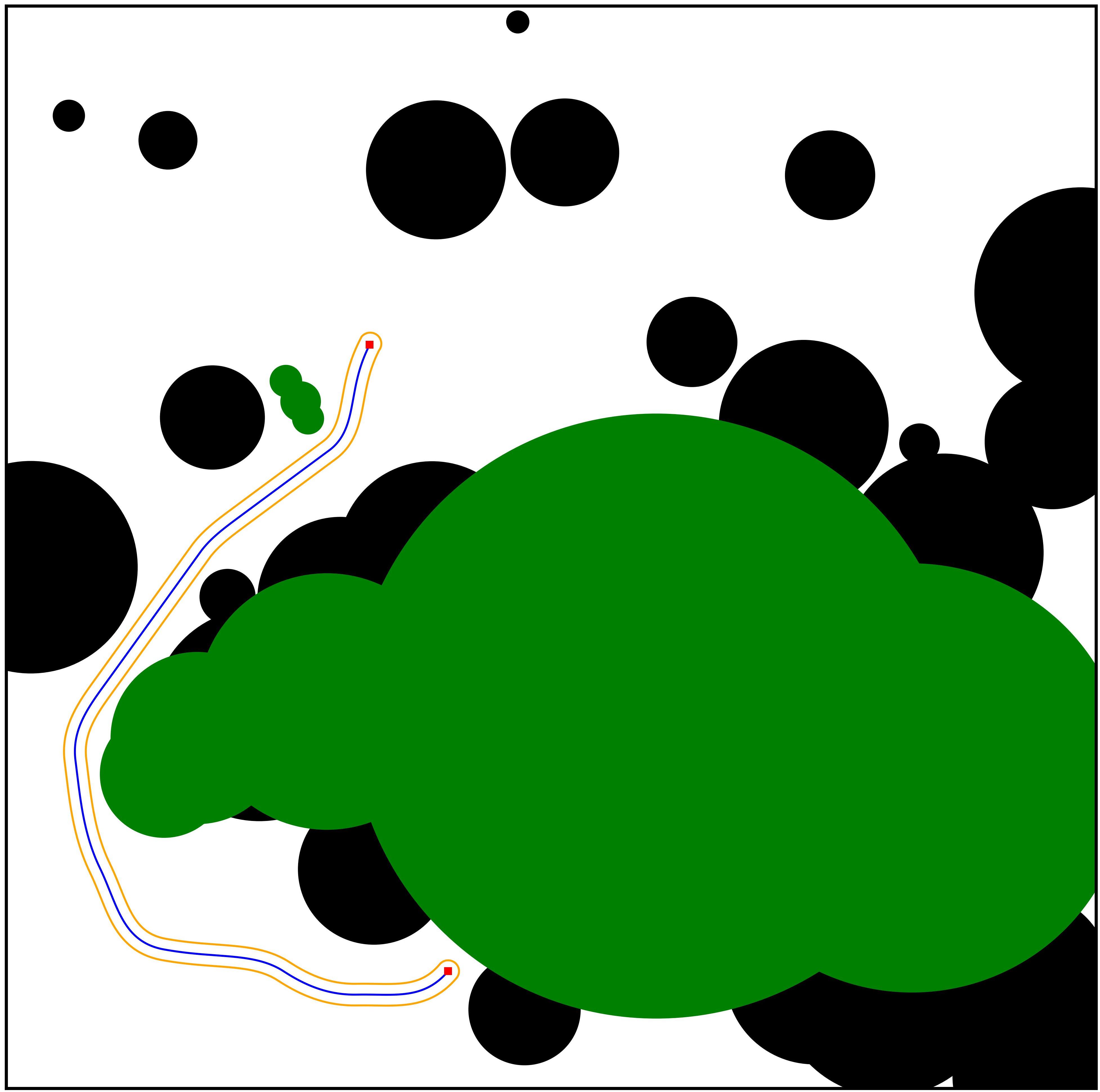}}
  	\subfloat[\label{fig.data.c1.2}]{
		\includegraphics[height=0.24\textwidth]{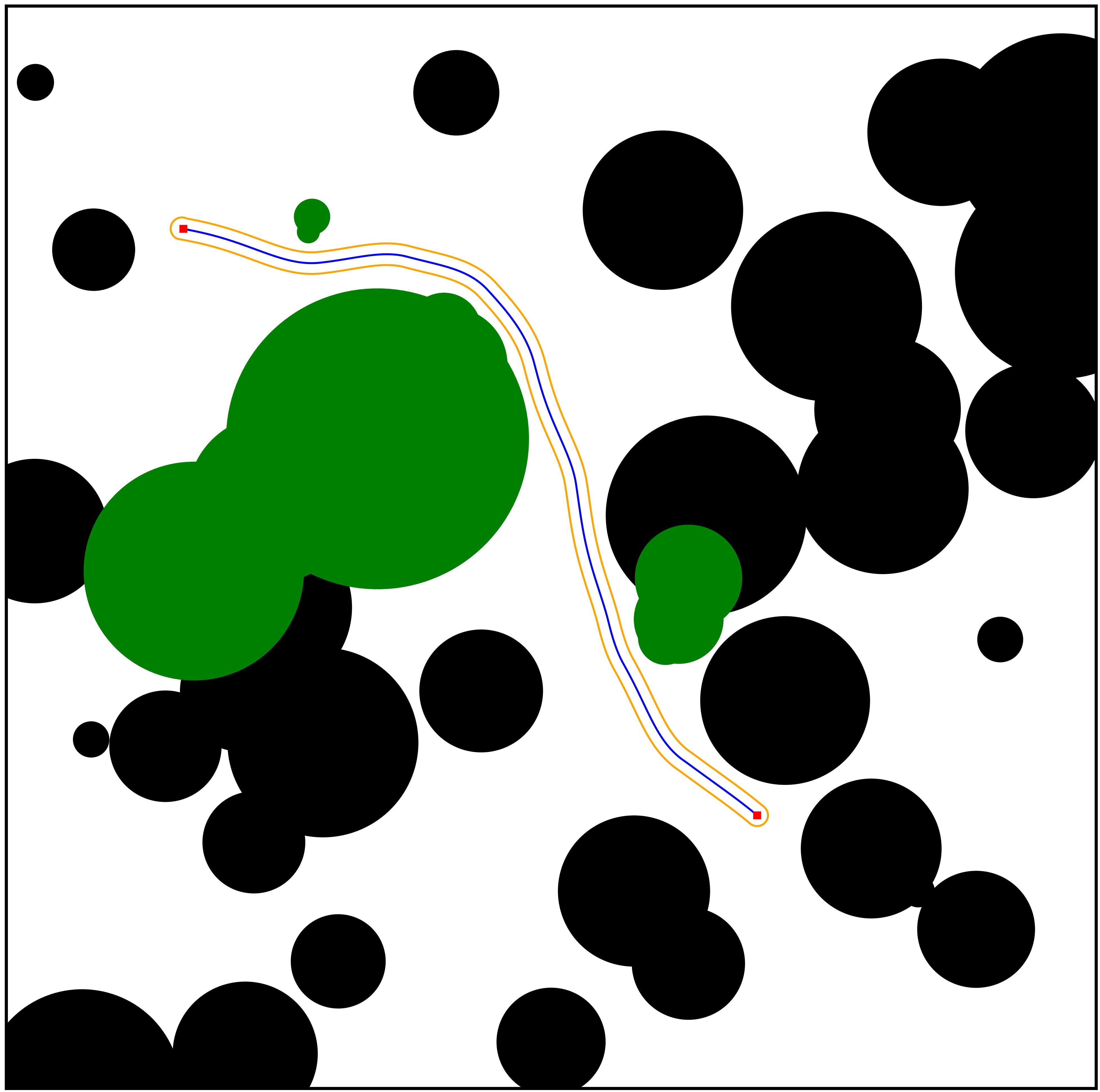}}
  	\subfloat[\label{fig.data.c1.3}]{
		\includegraphics[height=0.24\textwidth]{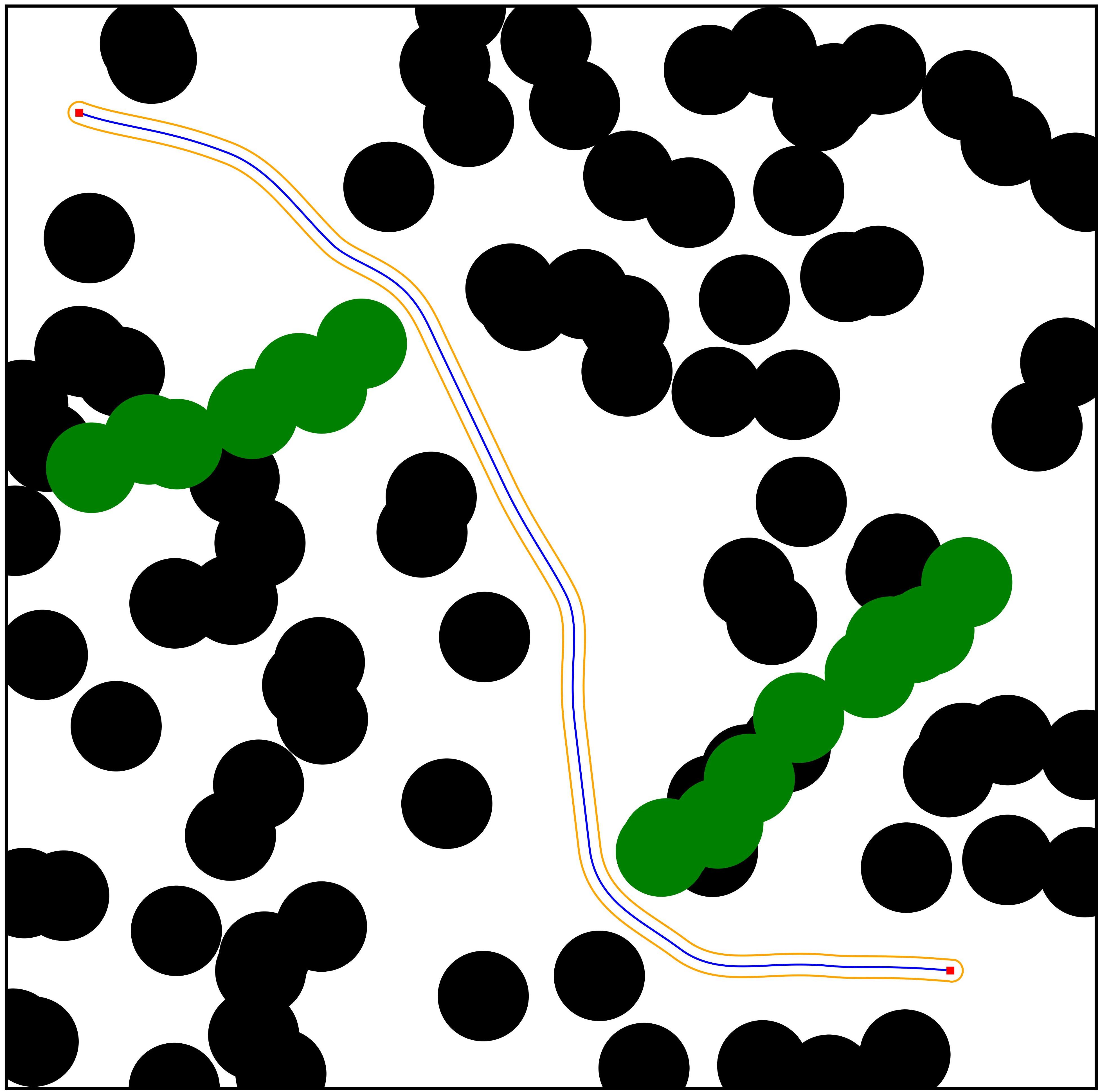}}
  	\subfloat[\label{fig.data.c1.4}]{
		\includegraphics[height=0.24\textwidth]{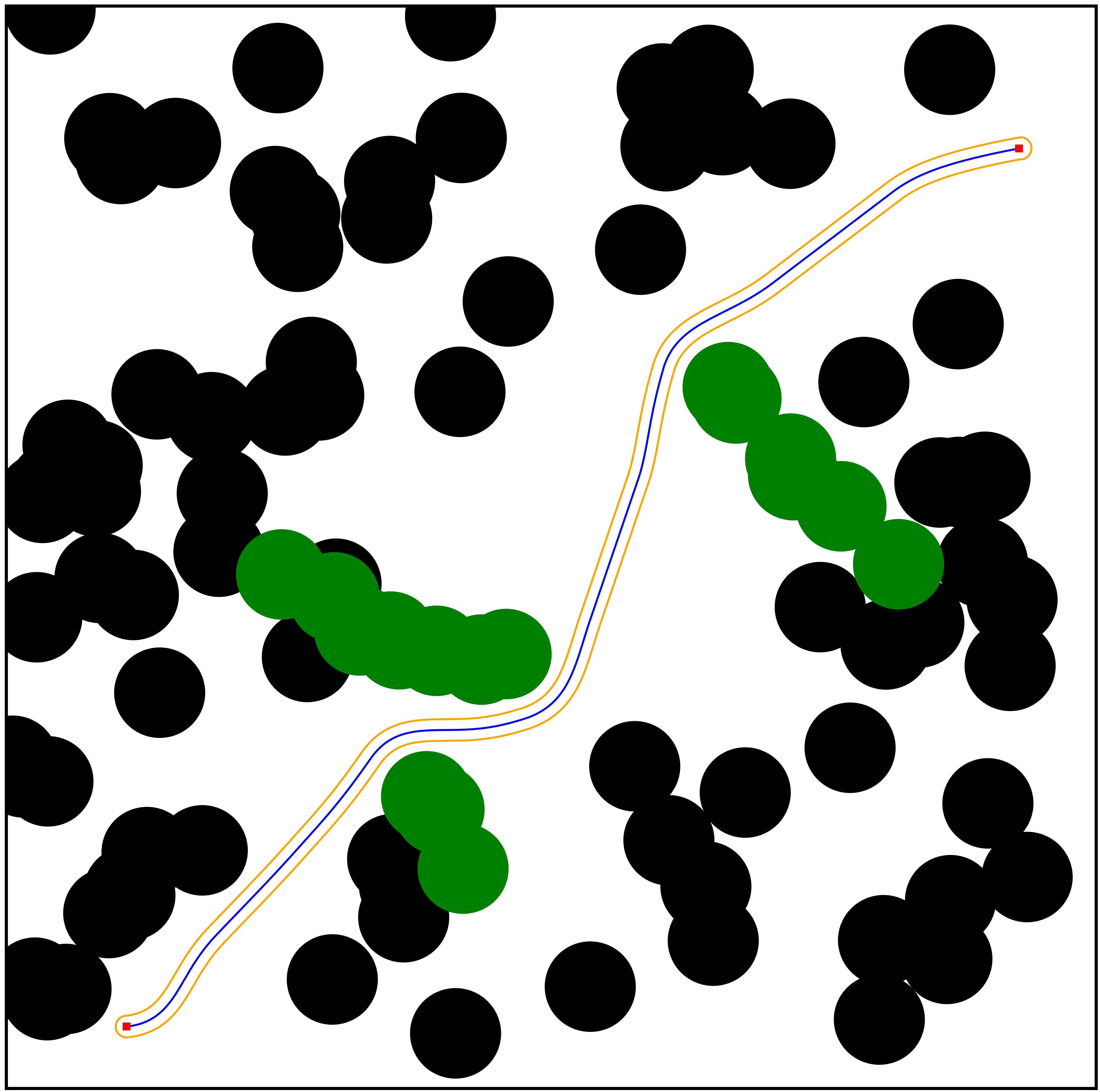}}
   \\
  	\subfloat[\label{fig.data.c3.1}]{
		\includegraphics[height=0.24\textwidth]{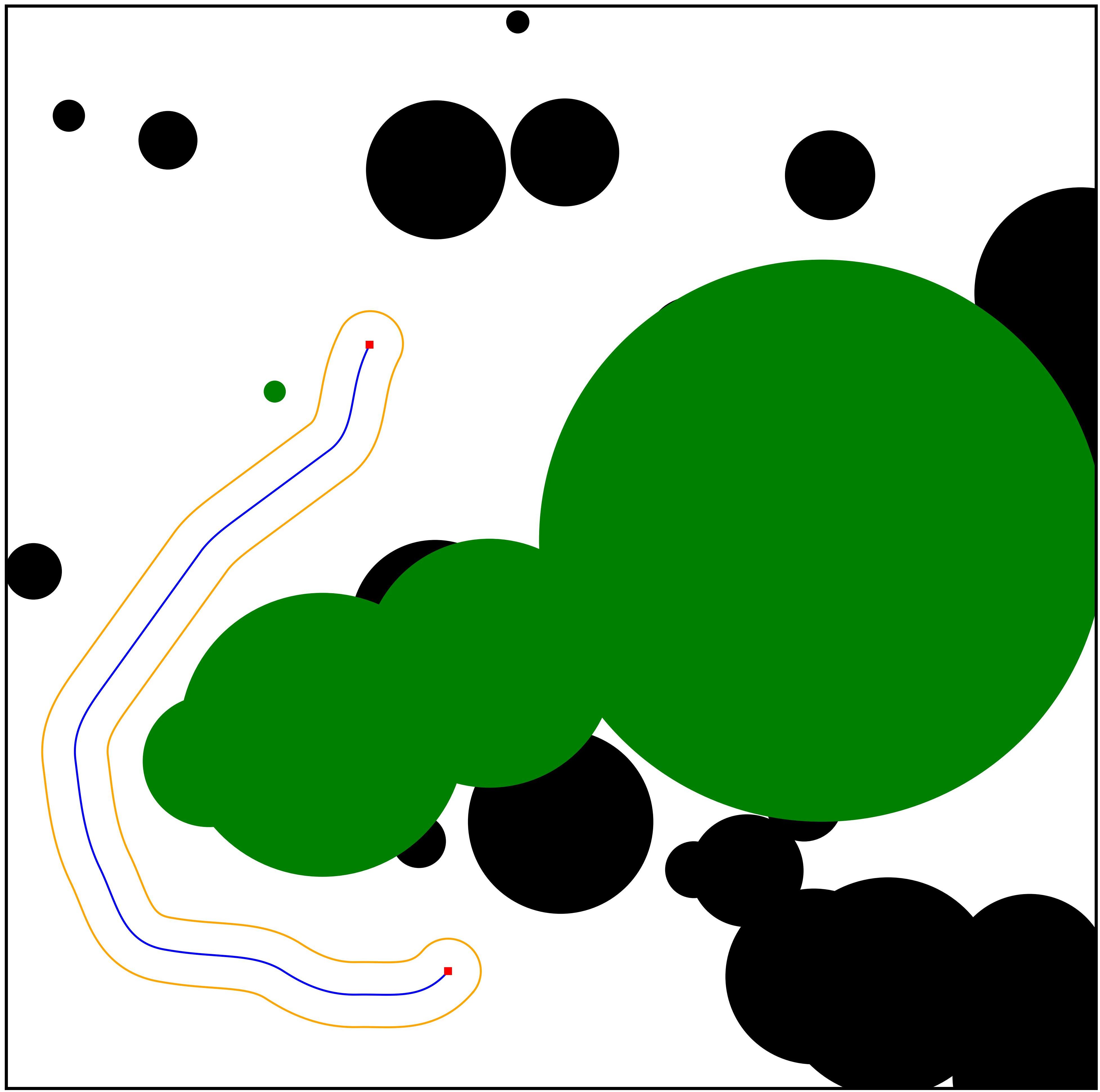}}
  	\subfloat[\label{fig.data.c3.2}]{
		\includegraphics[height=0.24\textwidth]{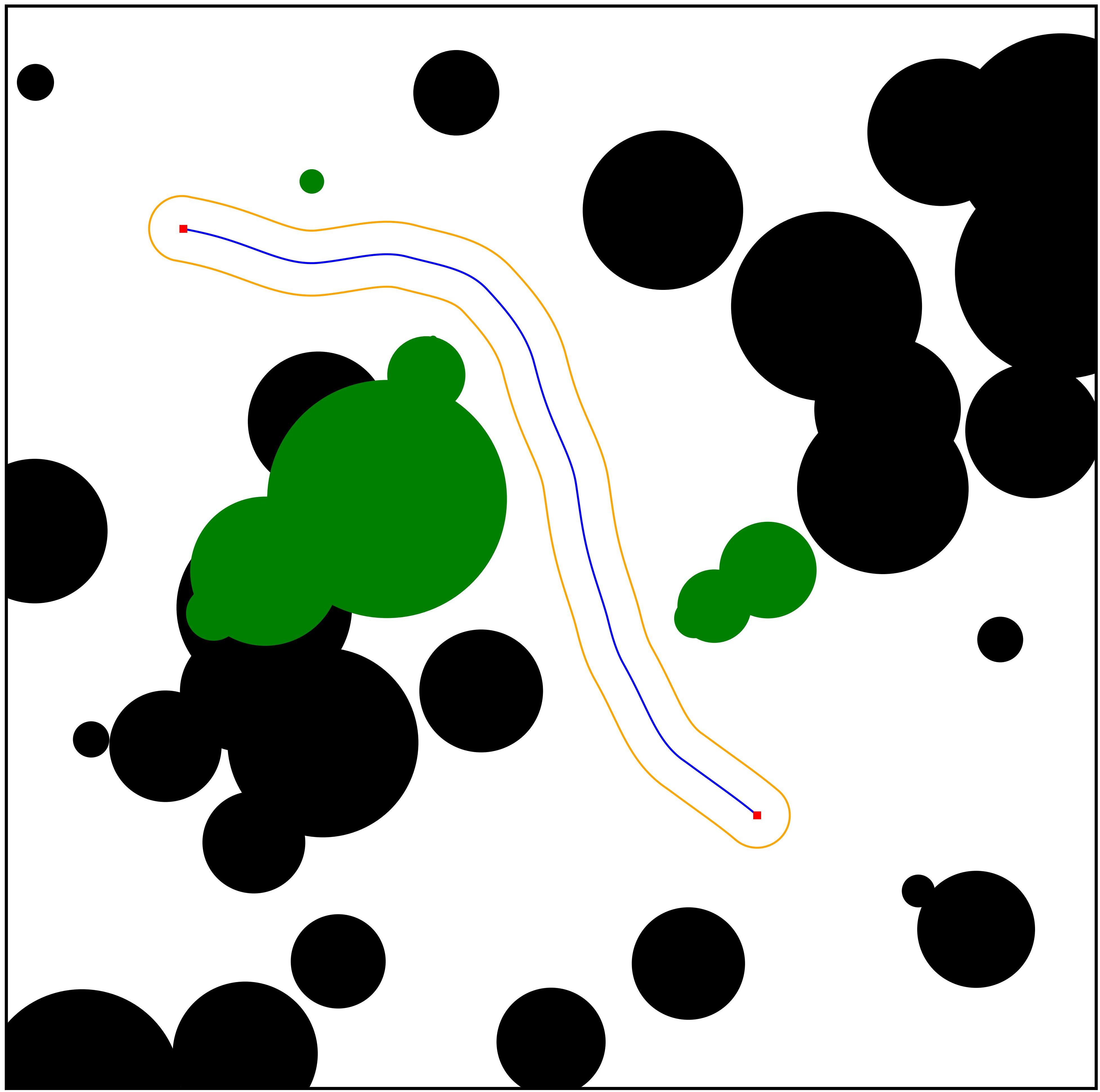}}
  	\subfloat[\label{fig.data.c3.3}]{
		\includegraphics[height=0.24\textwidth]{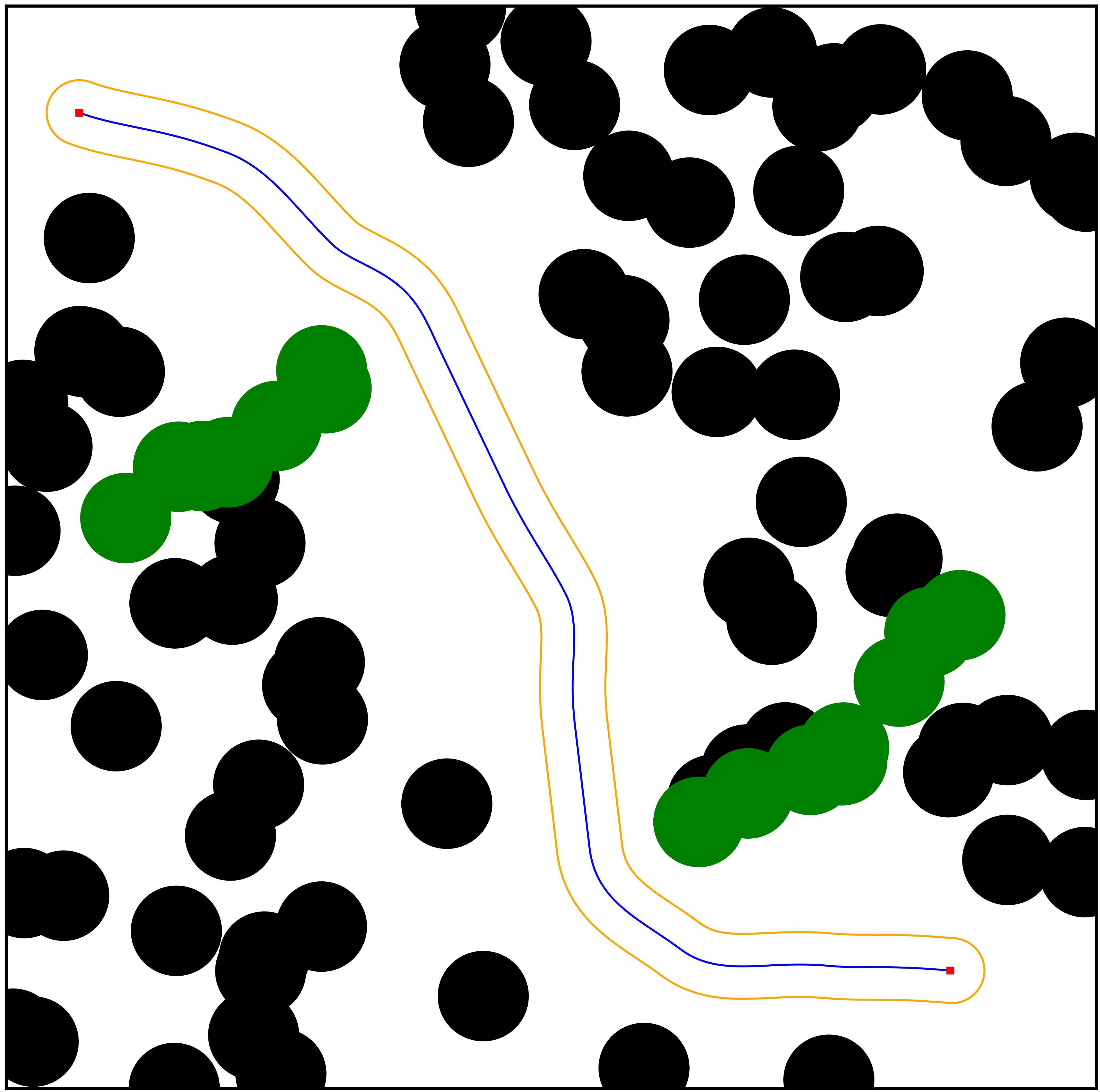}}
  	\subfloat[\label{fig.data.c3.4}]{
		\includegraphics[height=0.24\textwidth]{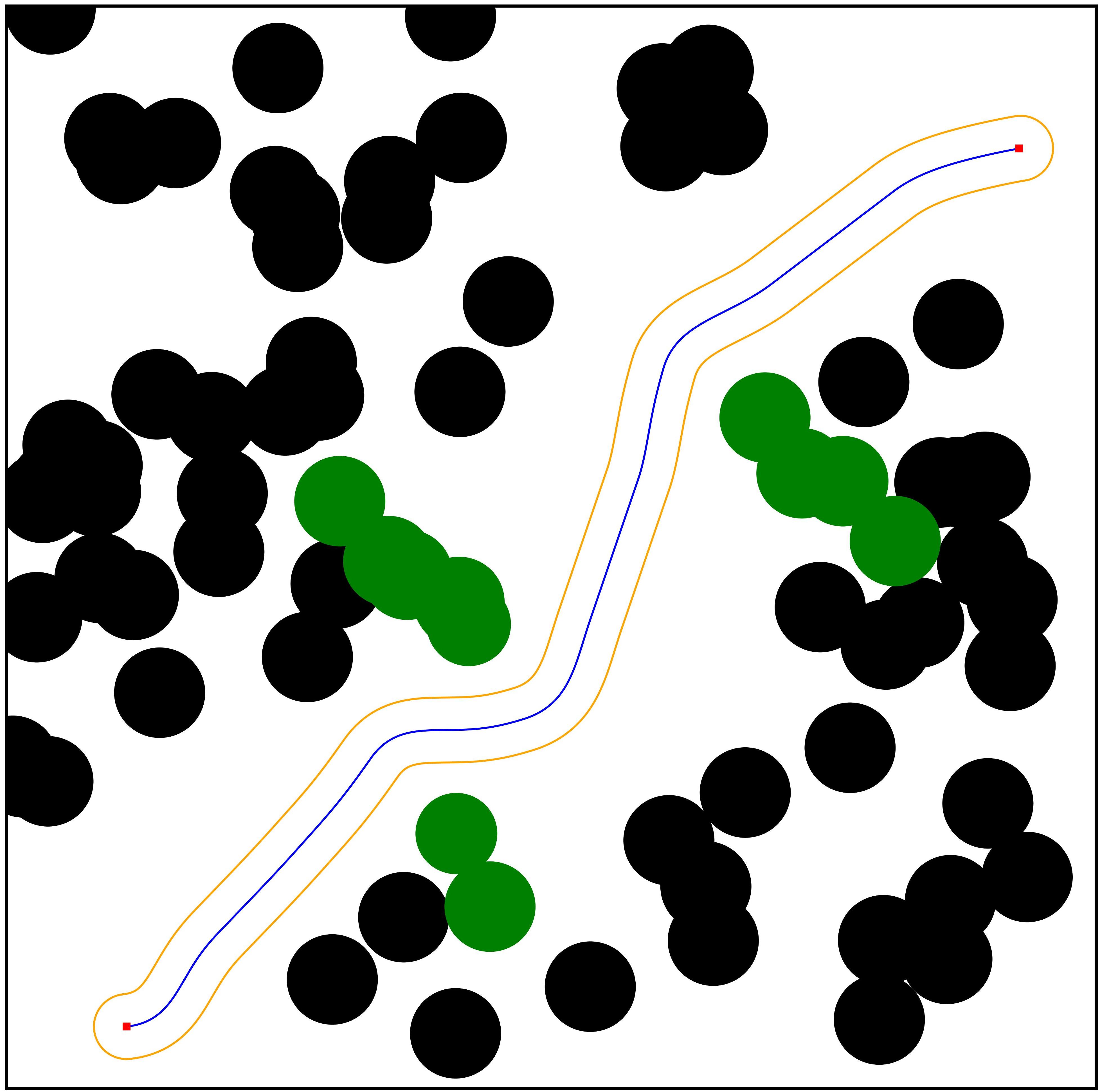}}
    \caption{Four examples of the data generated by EDaGe-PP with different clearance requirements. Each figure indicates the process of generating data, which can be divided into random path generation (blue), drawing the space of the path (orange), calculating the settings of the obstacles (green), and random obstacles in the state space (black). }
    \label{fig.data}
\end{figure*}

\subsection{Settings of the Obstacles}\label{sec:data:obs}
For making the path a near-optimal solution in any state space particular constraints need to be applied. In this paper, we use obstacles that are circular as the constraints for the path. The calculation of the obstacles' settings is presented in \algo{alg:obs}. The process of calculating the settings of obstacles is divided into two steps. The first step is to calculate the convex hull of the space and extract the subspace of the space. The second step is to calculate the size and position of obstacles in the subspace.
\subsubsection{Find Subspaces (\algolines{alg:obs}{alg:obs:hull}{alg:obs:sub})}
\par The collection of subspaces $S$ is defined as (\algoline{alg:obs}{alg:obs:sub}):
\begin{gather}
S=\left [ s^{(1)}, s^{(2)},\cdots, s^{(m)}, \right ],\\
s^{(k)}=\left [ \vec{s}_{1}^{(k)}, \vec{s}_{2}^{(k)},\cdots, \vec{s}_{w_{k}}^{(k)} \right ], k=1,2, \cdots, n_{1}, 
\end{gather}
where $m$ denotes the number of the subspaces, and $w_{k}$ denotes the number of the boundary's points of each subspace.
\par For each subspace, due to the path's waypoints being sampled with equal distance, the subspace can be found by the distance of the adjoining points of the convex hull. The indices of the pair of the adjoining points that the distance between them exceeds the threshold is defined as $f$, $g$. The relationship between $f$ and $g$ is described as:
\begin{gather}
g=\left\{\begin{matrix}
 0, f = m,\\
f + 1, f \neq m.
\end{matrix}\right.
\end{gather}
With the pair of adjoining points, the corresponding points in the list of the waypoints can be found, and the indices of the waypoints are denoted as $u$, $v$. The boundary of the subspace is the points between $\vec{p}_{u}^{'}$ and $\vec{p}_{v}^{'}$. 
\par With this rule, the volume threshold of the subspace is determined by the distance of the adjoining points as a hyper-parameter.
\subsubsection{Parameters of each Subspace (\algolines{alg:obs}{alg:obs:4}{alg:obs:6})}
In this section, we describe the algorithm on one subspace, for simplicity, we define the subspace as:
\begin{gather}
s=\left [ \vec{s}_{1}, \vec{s}_{2},\cdots, \vec{s}_{w} \right ],
\end{gather}
where $w$ denotes the number of the boundary's points of one subspace.
the directions are defined as (\algoline{alg:obs}{alg:obs:5}):
\begin{gather}
\Vec{z}=\Vec{h}_{f}-\Vec{h}_{g},\\
\Vec{n}=\left [\Vec{z}\left [1 \right ], -\Vec{z}\left [0 \right ] \right ].
\end{gather}
And the width of subspace in the direction of $\Vec{n}$ is defined as (\algoline{alg:obs}{alg:obs:6}):
\begin{gather}
w=\mathop{\mathrm{max}}\limits_{1\le i\le w}\ \left | \left ( \vec{s}_{i} - \Vec{h}_{r/s}\right)\cdot\Vec{n} \right |.
\end{gather}
\subsubsection{Settings of Obstacles (\algolines{alg:obs}{alg:obs:7}{alg:obs:17})}
\par For each subspace, the obstacles are set in the direction of $\vec{n}$ with a random translation that doesn't exceed the radius of each obstacle (\algolines{alg:obs}{alg:obs:8}{alg:obs:14}). The size of obstacles is random in the range of the twice width of each subspace. If the random obstacle is collision-free, the obstacle will insert the queue of obstacles of the subspace (\algolines{alg:obs}{alg:obs:15}{alg:obs:17}). The number of obstacles is not predetermined and because of the random size of each obstacle, the generation of obstacles will stop when the width of all obstacles exceeds the threshold which is the twice width of each subspace. 

\subsection{Random State Space}\label{sec:data:problem}
With the mask of the path's space and corresponding settings of obstacles, we can generate the path planning problems that the target path is a near-optimal solution. The Algorithm for generating Random State Space is presented in \algo{alg:prob}.
\par The translation and rotation of the path are generated randomly. Then the same translation and rotation are applied to the convex hull of the path's waypoints. Whether the convex hull of the path exceeds the boundary of the state space $\mathscr{R}$ is checked. If the convex hull is in the state space, the algorithm will continue executing, or (\algolines{alg:prob}{alg:prob:3}{alg:prob:6}) will be continually repeated until the success of boundary check or the times of repetition exceeds the predetermined threshold. When the times of repetition have exceeded the threshold, the algorithm of path generation will be judged as a failure and the algorithm will be executed from the start (\algoline{alg:data}{alg:data:1}). After the boundary check, the size and the position of the circle obstacles are randomly generated and only the collision-free obstacles will remain (\algolines{alg:prob}{alg:prob:8}{alg:prob:10}). The settings of obstacles of the path are placed with the same translation and rotation as the path. The original mask representing the waypoints of the path, and the original mask representing the path's space are doing the same translation and rotation with the path (\algolines{alg:prob}{alg:prob:11}{alg:prob:16}). The image of the state space $I_{Prob}$ and the mask of the path's space $I_{S}$ do Hadamard's product for making sure that no obstacles in the path's space, which means the path is collision-free. The calculation is described as:
\begin{gather}
I_{Prob} = \left [ a_{ij}^{(Prob)}\right ],\\
I_{S} = \left [ a_{ij}^{(S)}\right ],\\
I_{Prob} = I_{Prob} \odot I_{S} = \left [ a_{ij}^{(Prob)}a_{ij}^{(S)}\right ].
\end{gather}
\par At last, the data for training of PPNet is comprised of the mask of waypoints $I_{P}$, the mask of the path's space $I_{S}$, and the image of the state space $I_{Prob}$.

\section{PPNet: End-to-End Path Planner}\label{sec:net}
\begin{figure*}[!t]
    \centering
	   \includegraphics[width=0.98\textwidth]{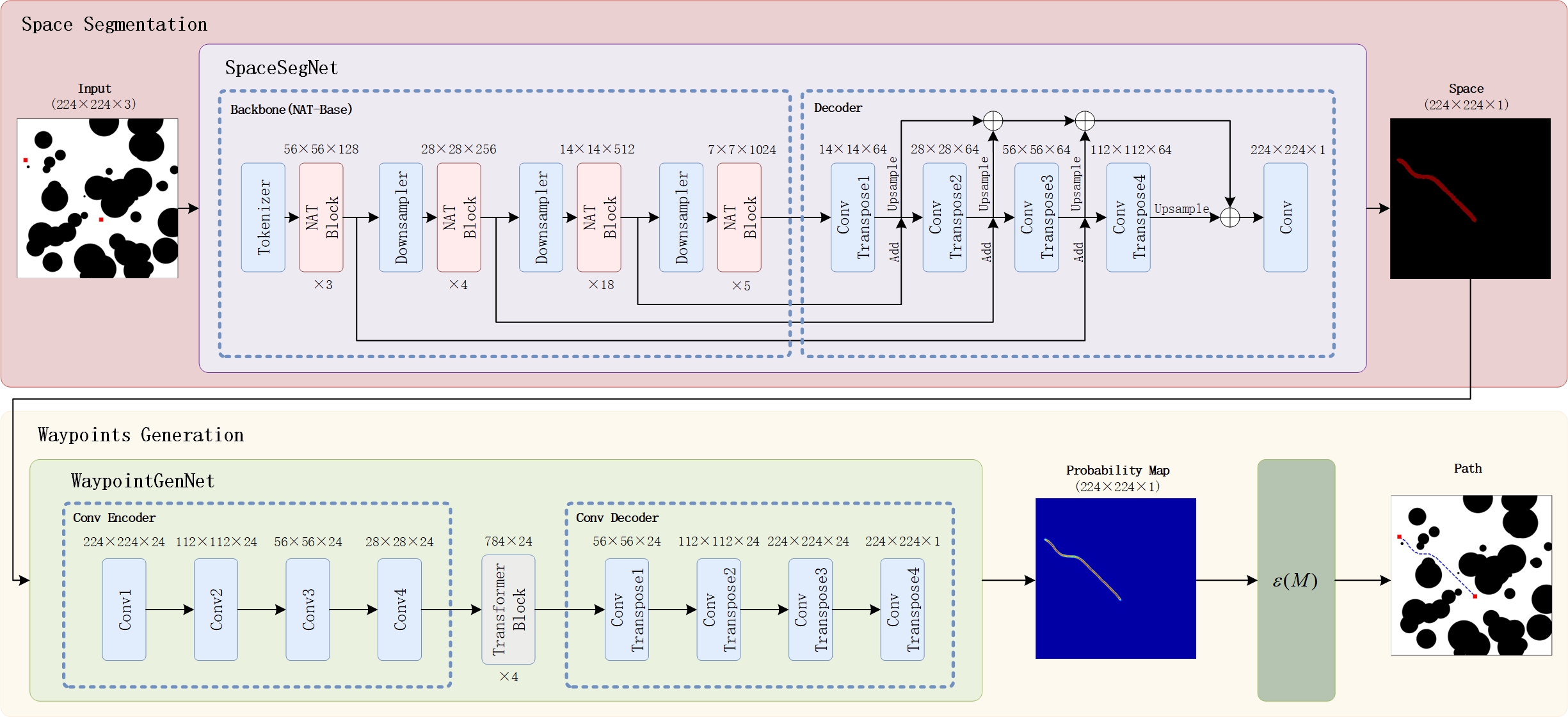}
    \caption{Model overview. PPNet consists of SpaceSegNet and WaypointGenNet. SpaceSegNet can solve the first subproblem, which is space segmentation. WaypointGenNet can solve the second subproblem, which is waypoints generation. }
    \label{fig.model}
\end{figure*}

In this section, we describe the two-stage structure of PPNet. To develop a method that can find a solution for end-to-end path planning while satisfying the clearance requirement, we found a key characteristic of solutions satisfying clearance requirements, which is there is a specific relationship between the path and its space, which means the path can be obtained by the shape of its space. Moreover, while the path can satisfy the clearance requirement, the space of the path should be completely in the free space. Therefore, we divide the path planning problem into two subproblems. The first one is to segment the path's space where the path is in it while satisfying the clearance requirement. The second one is to generate the waypoints of the path based on the shape of the path's space segmented in the first subproblem. This leads the structure of PPNet to a two-stage structure in which each stage solves one of the two subproblems (Fig.~\ref{fig.model}).
\par In Section~\ref{sec:net:seg}, we introduce the first subproblem which is space segmentation, and the design of the corresponding first-stage model of PPNet named SpaceSegNet. In Section~\ref{sec:net:gen}, the second subproblem which is waypoints generation, and the design of the second-stage model of PPNet named WaypointGenNet are introduced.

\subsection{Space Segmentation}\label{sec:net:seg}
Space segmentation is the task of segmenting the space of the path for the corresponding path planning problem. The segmented path's space should be the space within the boundary of the path that is calculated in Section~\ref{sec:data:space}. This makes the shape of the space can be used for generating the corresponding path. Such a problem can be described as:
\begin{equation}
S=f_{S}\left ( x_{init}, x_{end},  \mathcal{O} \right ) ,
\end{equation}
where $S$ denotes the path's space, and $\mathcal{O}$ denotes the obstacles in the state space.
\par To make it easier to solve this subproblem by neural networks, we further transfer the problem into a binary classification task, which can be defined as:
\begin{equation}
y=f'_{S}\left ( x, x_{init}, x_{end},  \mathcal{O} \right ) ,
\end{equation}
where $y$ is a binary result of whether the $x$ is in the space of the path for the corresponding path planning problem, which is described by $x_{init}$, $x_{end}$,  and $\mathcal{O}$.
\par Due to the difficulty of convergence when neural networks take the continuous representation of $x_{init}$, $x_{end}$, and $\mathcal{O}$ as input, $x_{init}$, $x_{end}$, and $\mathcal{O}$ are discretized into a picture that uses markers to denote $x_{init}$, $x_{end}$. After discretization, the problem can be defined as:
\begin{equation}
Y=\mathcal{S}\left ( \mathcal{M} \right ) ,
\label{epu.prob1.2class}
\end{equation}
where $\mathcal{S}$ denotes the function of SpaceSegNet, $\mathcal{M}$ denotes the picture that describes the path planning problems, and $Y$ denotes the result of space segmentation.
\par This leads the problem to a segmentation task, so we develop a segmentation model to solve the first subproblem which is space segmentation, named SpaceSegNet. SpaceSegNet takes the picture that describes the path planning problem as the input and outputs the mask of the path's space. The output of SpaceSegNet is a mask of the path's space which is a binary picture that indicates whether the point is in the path's space, which means each pixel is the result of the problem defined in Equation~\ref{epu.prob1.2class} of the corresponding point $x$. And we use such a picture as the representation of the path's space.

\subsection{Waypoints Generation}\label{sec:net:gen}
Waypoints generation is the task of generating the waypoints by the shape of the path's space. The problem can be described as:
\begin{gather}
W=f_{G}\left ( S \right ) ,
\end{gather}
where $W$ denotes the waypoints of the path.
\par We transfer the problem into a two-step process, which is defined as:
\begin{gather}
M=\mathcal{G}\left ( Y \right ) ,\\
P^{*}=\mathcal{E}\left ( M \right ) ,
\end{gather}
where $M$ denotes the probability map of whether the point $x$ is the waypoint of the path, and $P^{*}$ denotes the waypoints of the path.
\par The first step can be seen as a generation task, so we develop a generation model to solve it, named WaypointGenNet. WaypointGenNet is designed as a Transformer-based Autoencoder. WaypointGenNet takes the binary picture $Y$ that represents the path's space as input and outputs the probability map of the waypoints.
\par The second step is to extract the waypoints from the probability map $M$. $\mathcal{E}\left ( \cdot \right )$ denotes the simple rule of extracting waypoints from the probability map that is generated by WaypointGenNet, which is defined as:
\begin{equation}
x_{next} =  \underset{X_{neighbor}}{{\arg\max} \, M},
\end{equation}
where $X_{neighbor}$ is the neighborhood points of the current point on the probability map, which starts at the initial point.
The probability map $M$ can be seen as the solution to the given path planning problem. The computation of extracting the waypoints from the probability map can be finished during the executed time. So the total computation time of our method is defined as:
\begin{equation}
t =  t_{\mathcal{S}} + t_{\mathcal{G}} ,
\end{equation}
where $t_{\mathcal{S}}$ is the inference time of SpaceSegNet $\mathcal{S}$, $t_{\mathcal{G}}$ is the inference time of WaypointGenNet $\mathcal{G}$.

\section{Experiments}\label{sec:exp}
In this section, we demonstrate the performance of PPNet through several simulations. The system used for experiments has a 3.20 GHz× 8 AMD Ryzen 7 5800H processor with 16-GB RAM and NVIDIA GeForce RTX 3060 GPU.
\par For training, we adopt a polynomial learning rate decay schedule and employ SGD as the optimizer. Momentum and weight decay are set to 0.9 and 0 respectively for all the experiments on the three datasets. We set an initial learning rate of 0.08 for all experiments. We set the batch size to 8 and the total iteration to 80,000 for experiments of comparison of different encoders of SpaceSegNet. And we used mmsegmentation~\cite{mmseg} for building SpaceSegNet and experiments in Section~\ref{sec:exp:str}. The path planning problems used in this section are generated by EDaGe-PP, which contains up to 50 obstacles in each problem. In the experiments of previous learning-based methods, they test the methods only in a few maps, eg. NRRT* in 4 maps, and MPNet in 10 maps. Such a setting is reasonable for classic methods, but learning-based methods need sufficient different maps to validate the generalization. Therefore, we evaluate our methods on a dataset that each data is a path planning problem in a different map, which can validate the generalization of our method. Unlike the previous works combining machine learning techniques and sampling-based methods, our method is a deep learning method without implementing traditional path planning methods, and our method can only solve path planning problems in 2D scenarios. Developing methods for solving path planning problems in higher dimension space in an end-to-end way will be our future work.

\subsection{Comparison of Different Structures of SpaceSegNet}\label{sec:exp:str}
In this section, we test the performance of different structures of the segmentation model in path planning. For evaluating different structures of SpaceSegNet, the size of the dataset used in this section is 320,000, which means the dataset contains 320,000 different maps. Unlike the traditional segmentation tasks that can segment target regions by texture in images, eg. ADE20K, and crack segmentation, the space of the target path should be completely in the free space, which means the target region is in the white region has no texture, this makes the task much more difficult.
\subsubsection{Comparison of Encoders}
\par In the comparison of encoders, four kinds of variants of vision Transformer which are Vision Transformer (ViT)~\cite{vit}, Swin Transformer (Swin)~\cite{swin}, Neighborhood Attention Transformer (NAT)~\cite{nat} and Dilated NAT(DiNAT)~\cite{dinat}, are used. For making all models with different encoder parameters similar in number, the base variants of ViT, Swin, and NAT are used in the experiment of comparison, which are ViT-Base, Swin-Base, NAT-Base, and DiNAT-Base. Because ViT has no multi-level feature, we use the Progressing UPsampling head (PUP) of SEgementation TRansformer (SETR)~\cite{setr} as the decoder for all models in comparison of different encoders with the head of Fully Convolutional Networks (FCN) as the auxiliary head. 
\par In the experiments on Table~\ref{tab:encoder}, the reason that the performance of models with these encoders in path planning is quite different from the performance in classical CV tasks is the different attention mechanisms used in these encoders introduced in different assumptions, which have quite different influence in path planning.
\par \textbf{ViT} uses the vanilla Self-Attention (SA)~\cite{sa} that each patch can calculate attention with all patches directly. Nevertheless, this makes SA can be computationally expensive with the growth of the number of patches, SA introduces no assumption. 
\par \textbf{Swin} uses Shifted Window Self-Attention (SWSA)~\cite{swin}. SWSA makes the SA more computationally efficient by only calculating the attention between patches in the windows. The patches that are in the different windows establish correlations by the shifted windows. This makes some patches that are adjoining have no attention calculation directly which is unreasonable in path planning. In our experiments, the model using Swin as the encoder had even worse performance than the model using ViT, although Swin outperforms ViT in all classical tasks of Computer Vision (CV). 
\par \textbf{NAT} uses Neighborhood Attention (NA)~\cite{nat}. NA can be seen as a sliding window variant of SWSA. In NA, each patch only calculates attention with the adjacent patches which makes the computation more efficient, and the correlation of adjacent patches tends to be higher than SA. Such an assumption is reasonable for path planning. And in our experiment, the model using NAT as an encoder obtained the best performance. 
\par \textbf{DiNAT} uses Dilated NA (DiNA)~\cite{dinat} which is inspired by dilated convolution that can make the window of NAT extremely large with the same amount of computation. With the larger window, the model can obtain the receptive field as large as the task needs. But in DiNA, patches can not calculate attention directly with adjoining patches after the first layer which makes the model using DiNAT can not converge in path planning.
\par In summary, the model using NAT obtained the best performance due to the assumption which is introduced by NA. Nevertheless Swin outperforms ViT in all classical tasks in CV, and models using Swin and ViT perform similarly in path planning. Model using DiNAT can not converge in path planning.
\begin{table}[t]
    \caption{\textsc{Comparison of Different Encoders}}
    \centering
    \begin{tabular}{|c|c|c|c|}
        \hline
        Model& \# of Params     & FLOPs  & IoU(\%)\\
        \hline
        ViT  &  98.2 M &   57.23 G & 54.72\\
        Swin  &  102.1 M &   55.21 G & 53.67\\
        NAT &  104.1 M &   53.47 G & \textbf{61.89} \\
        DiNAT &  104.1 M &   57.62 G & 0 \\
        \hline
    \end{tabular}
\label{tab:encoder}
\end{table}

\begin{table}[t]
\caption{\textsc{Comparison of Different Decoders}}
    \centering
    \begin{tabular}{|c|c|c|c|}
        \hline
        Model& \# of Params     & FLOPs  & IoU(\%)\\
        \hline
        PUP-512  &  104.1 M &   53.47 G & 61.89\\
        UPerHead-512  &  123.2 M &   54.16 G & \textbf{64.15}\\
        UPerHead-256 &  100.1 M &   24.00 G & 63.91 \\
        UPerHead-128 &  93.4 M &   16.38 G & 63.41 \\
        UPerHead-64 &  91.2 M &   14.44 G & 63.53 \\
        \hline
    \end{tabular}
\label{tab:decoder}
\end{table}

\subsubsection{Comparison of Decoders}
\par In the comparison of decoders, due to the multi-scale feather that NAT extracts from input, the head of UPerNet (UPerHead)~\cite{upernet} is chosen to compare with PUP which is used in the comparison of different encoders. For making the number of parameters and FLOPs of NAT-PUP-512 and NAT-UPerHead-512 similar in number, the numbers of PUP's embedding dimensions and upsampling operations are set to 512 and 5 respectively. The other follows the configuration for SETR-PUP in SETR. The number of UPerNet's embedding dimensions is 512. The other follows the configuration for UPerNet-NAT in NAT. 
\par Following the configuration of decoders above, the number of parameters of NAT-PUP-512, and NAT-UPerHead-512 are 104.1M, and 123.2M respectively. The numbers of FLOPs of NAT-PUP-512, and NAT-UPerHead-512 are 53.47G, and 54.16G respectively. In the experiment on Table~\ref{tab:decoder}, all NAT-UPerHead models outperform NAT-PUP-512. As the performance of NAT-UPerHead models is quite similar, we choose the NAT-UPerHead-64 as the model for SpaceSegNet.

\begin{table*}[t]
    \caption{\textsc{Data Generation Time, Path Cost, and Success Rate Comparison of The Variants of PPNet }}
    \centering 
    \begin{threeparttable}
    \begin{tabular}{|c|c|c|c|}\hline
    Methods&Generation Time&Path cost&Success rate$(\%)$\\\hline
        RRT* &25.6min (+14.2h) &$\mathbf{37.3 \pm 9.44}$&$40.80$ \\
        IRRT* & 10.1min (+14.1h) &$37.8 \pm 8.20$&$48.84$ \\
        BIT* &5.96min (+16.3h)  &$37.8 \pm 8.19$&$45.21$ \\
        ABIT* &5.86min (+16.0h)  &$37.9 \pm 8.04$&$40.92$ \\\hline
        EDaGe-PP &\textbf{0.963min (+0.425h)}  &$39.0 \pm 7.39$&$\mathbf{94.47}$  \\\hline
    \end{tabular}
    \begin{tablenotes}
    \footnotesize
    \item[*]The variants of PPNet are trained by datasets which were generated by EDaGe-PP and data generation methods that solutions were found by RRT*, IRRT*, BIT*, and ABIT* respectively. 
    \item[**]Data generation time is divided into two parts, which are time for generating solutions and time for generating data with the format for training of PPNet.   
      \end{tablenotes}
    \end{threeparttable}
    \label{tab:data}
\end{table*}

\begin{figure*}[!t]
    \centering
	\subfloat[\label{fig.c1.1}]{
		\includegraphics[height=0.24\textwidth]{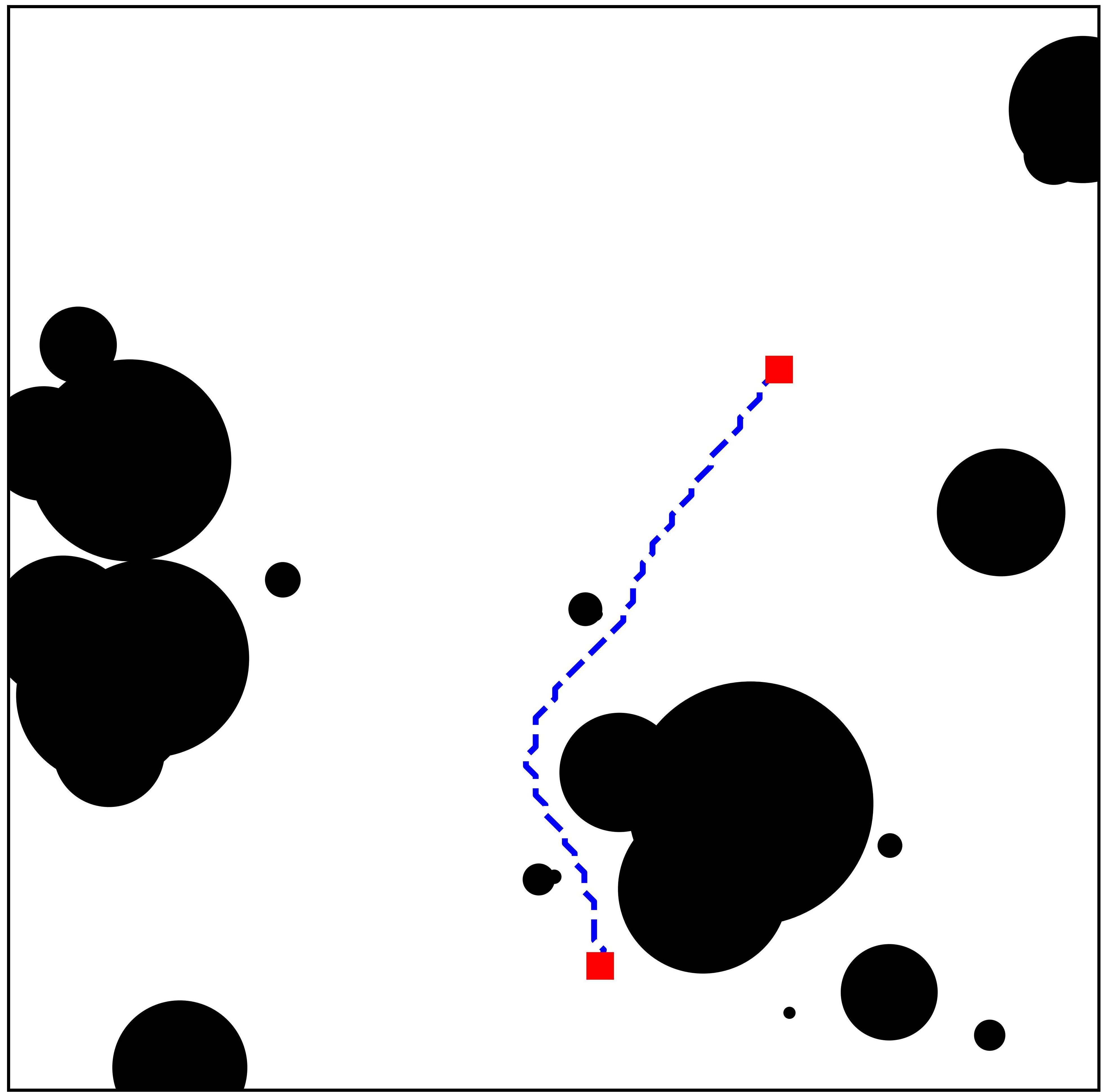}}
  	\subfloat[\label{fig.c1.2}]{
		\includegraphics[height=0.24\textwidth]{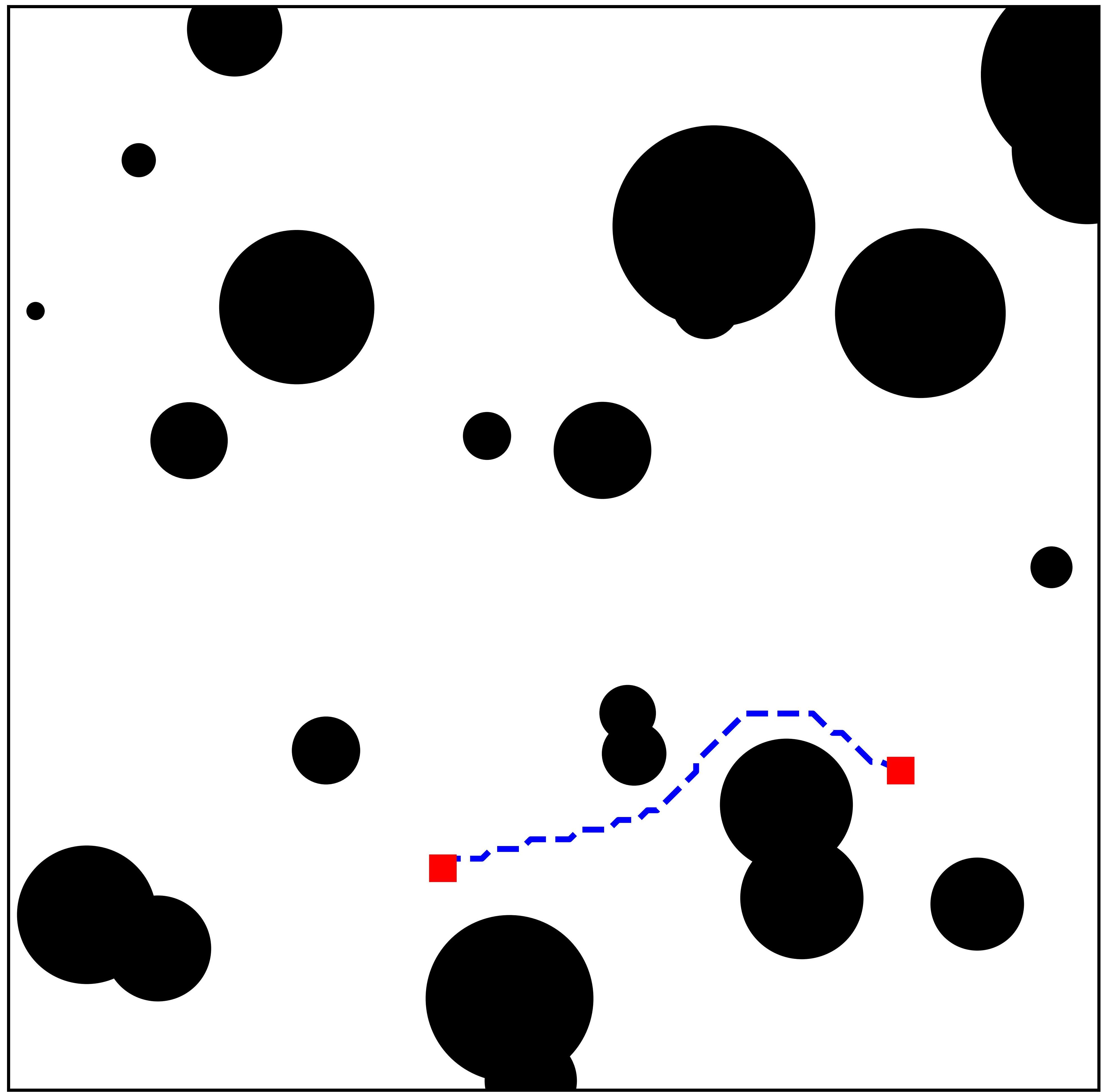}}
    \subfloat[\label{fig.c1.3}]{
		\includegraphics[height=0.24\textwidth]{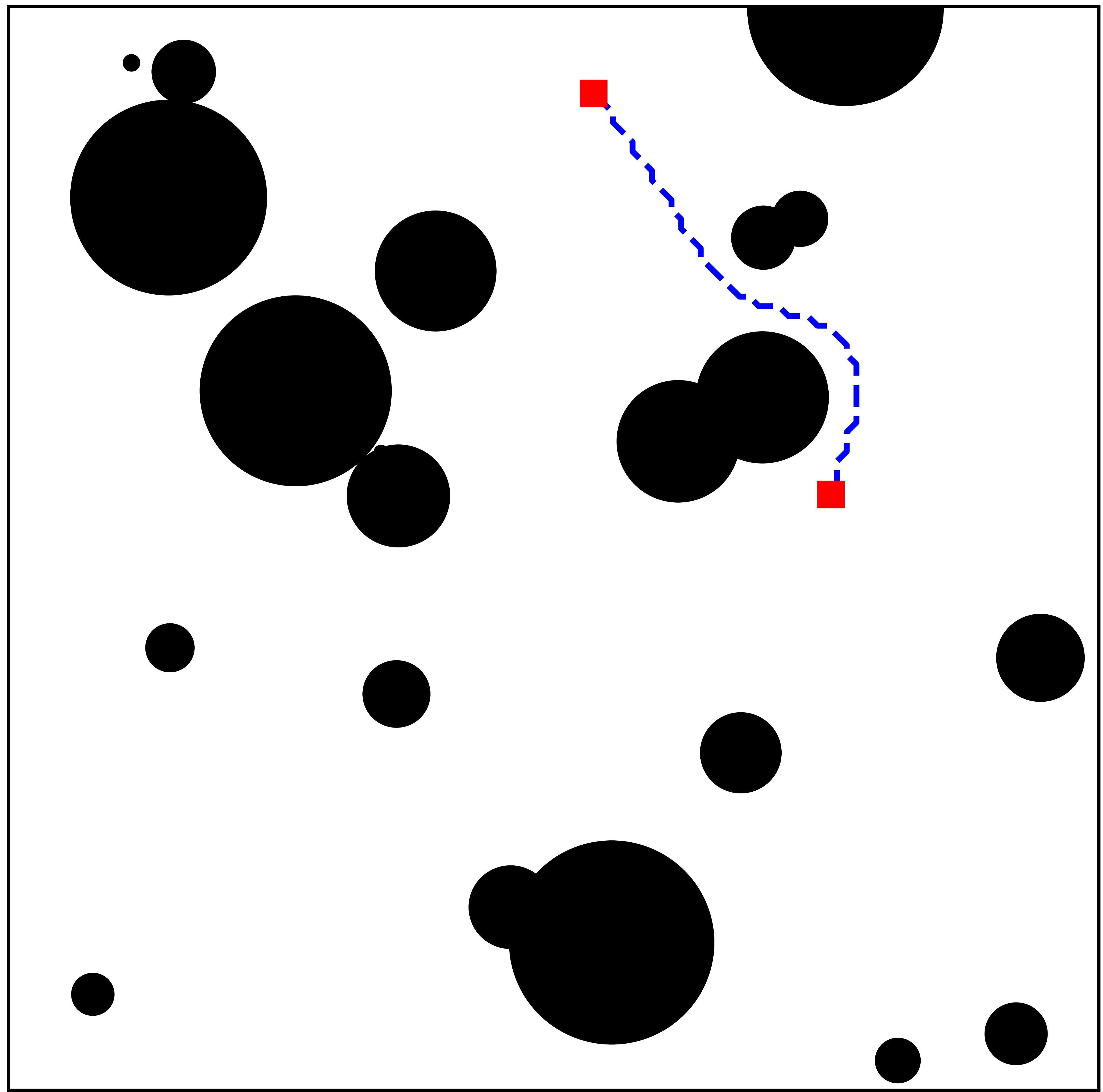}}
    \subfloat[\label{fig.c1.4}]{
		\includegraphics[height=0.24\textwidth]{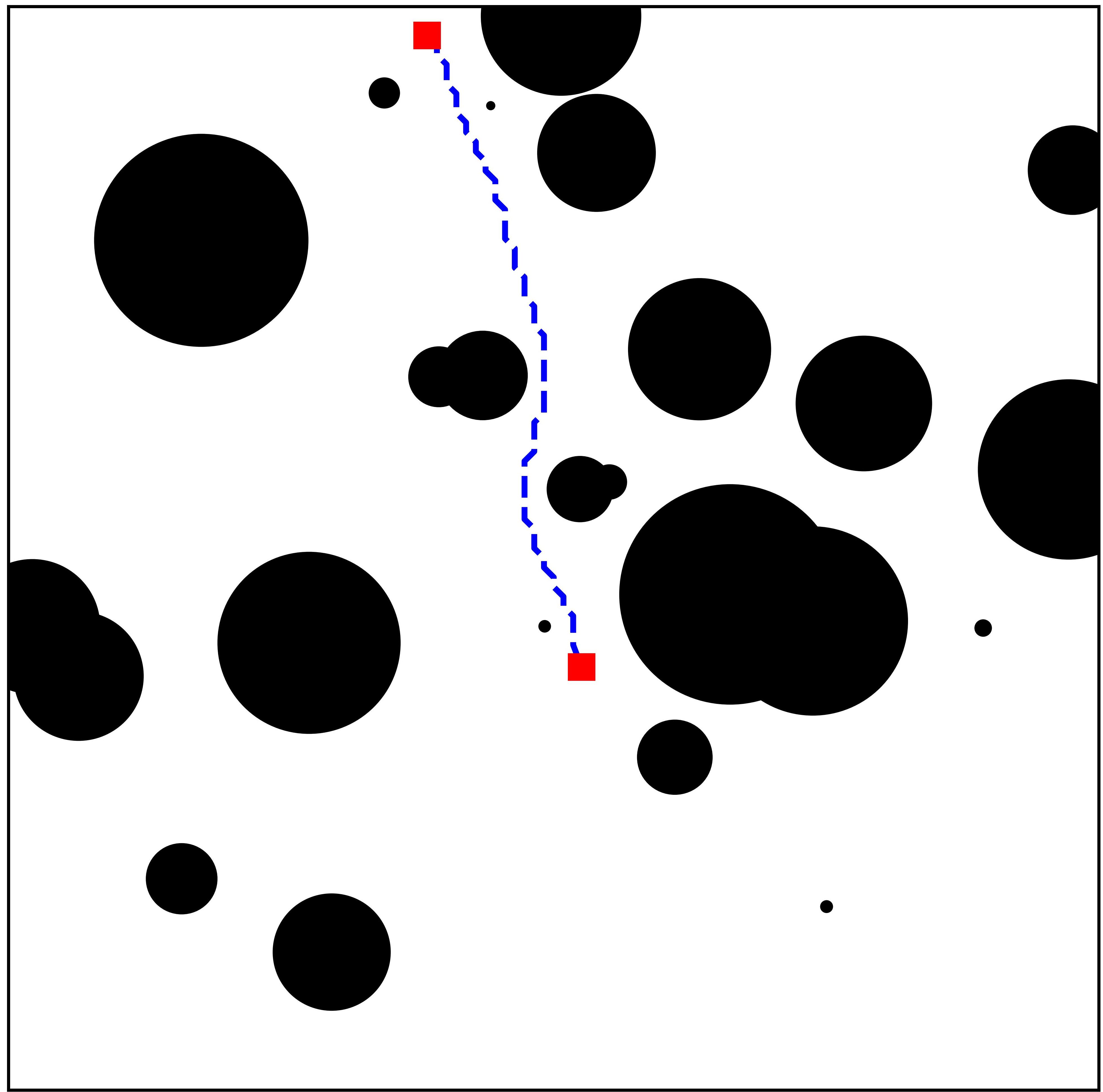}}
    \\
    \subfloat[\label{fig.c3.1}]{
		\includegraphics[height=0.24\textwidth]{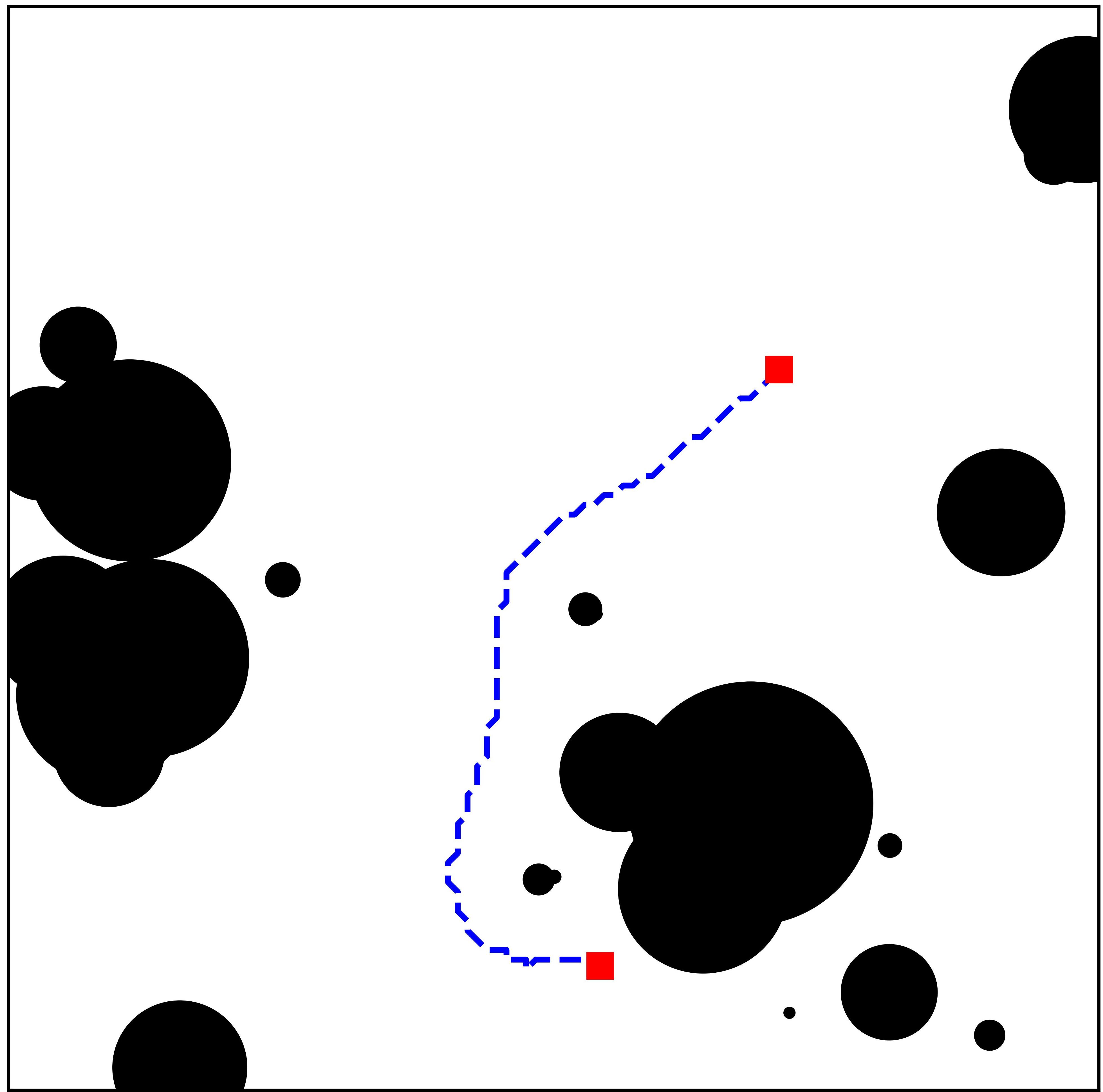}}
    \subfloat[\label{fig.c3.2}]{
		\includegraphics[height=0.24\textwidth]{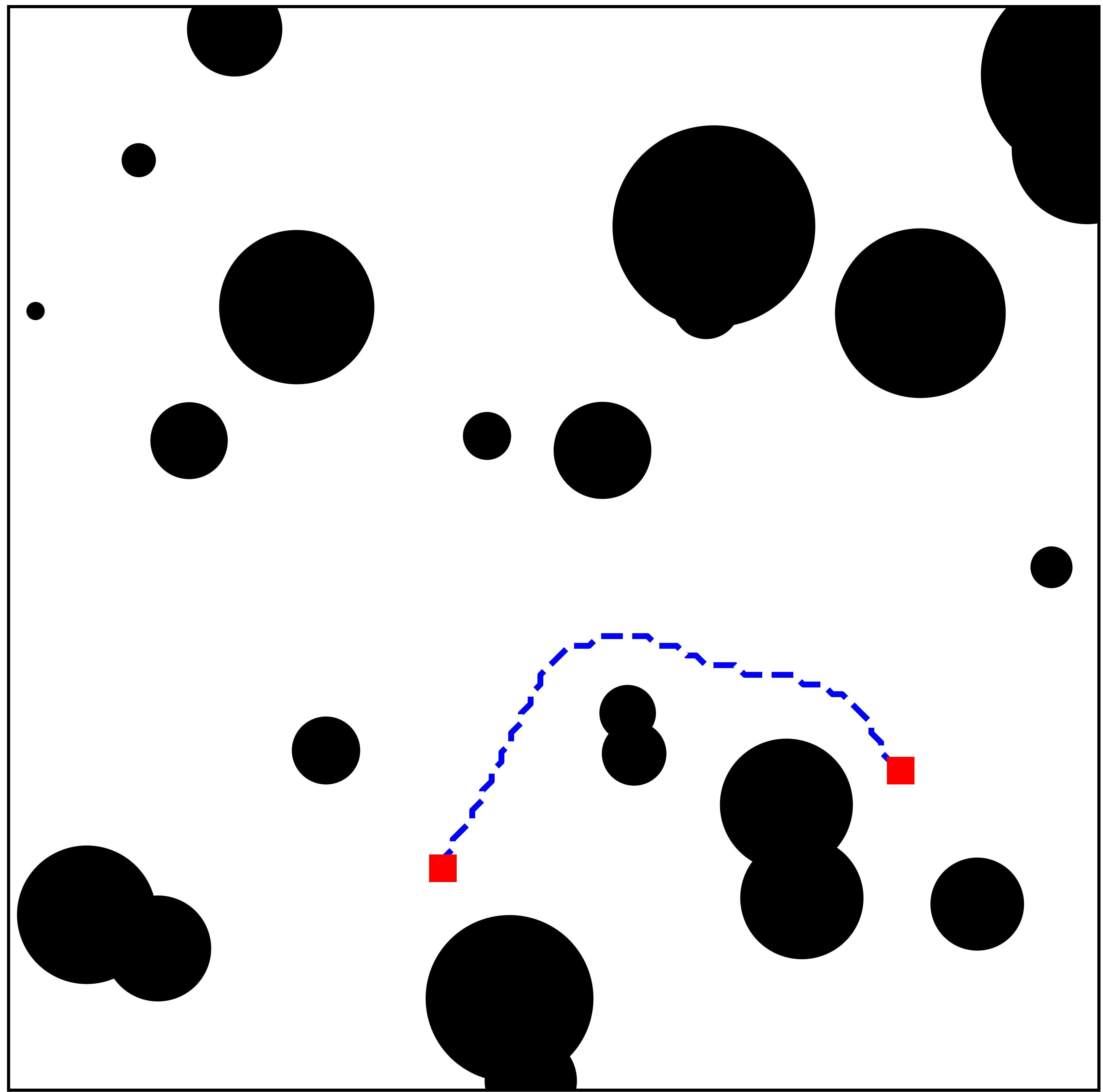}}
    \subfloat[\label{fig.c3.3}]{
		\includegraphics[height=0.24\textwidth]{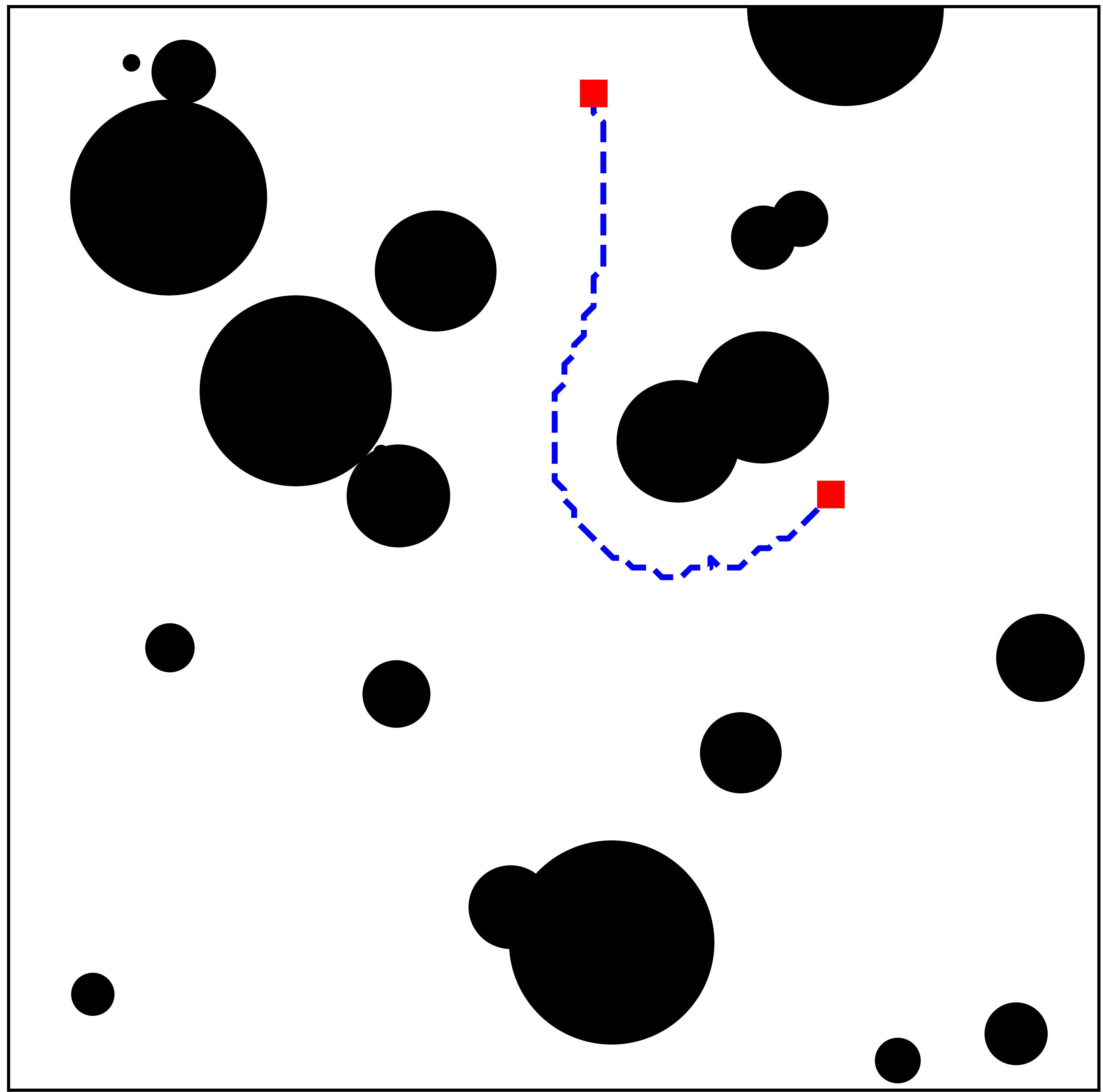}}
    \subfloat[\label{fig.c3.4}]{
		\includegraphics[height=0.24\textwidth]{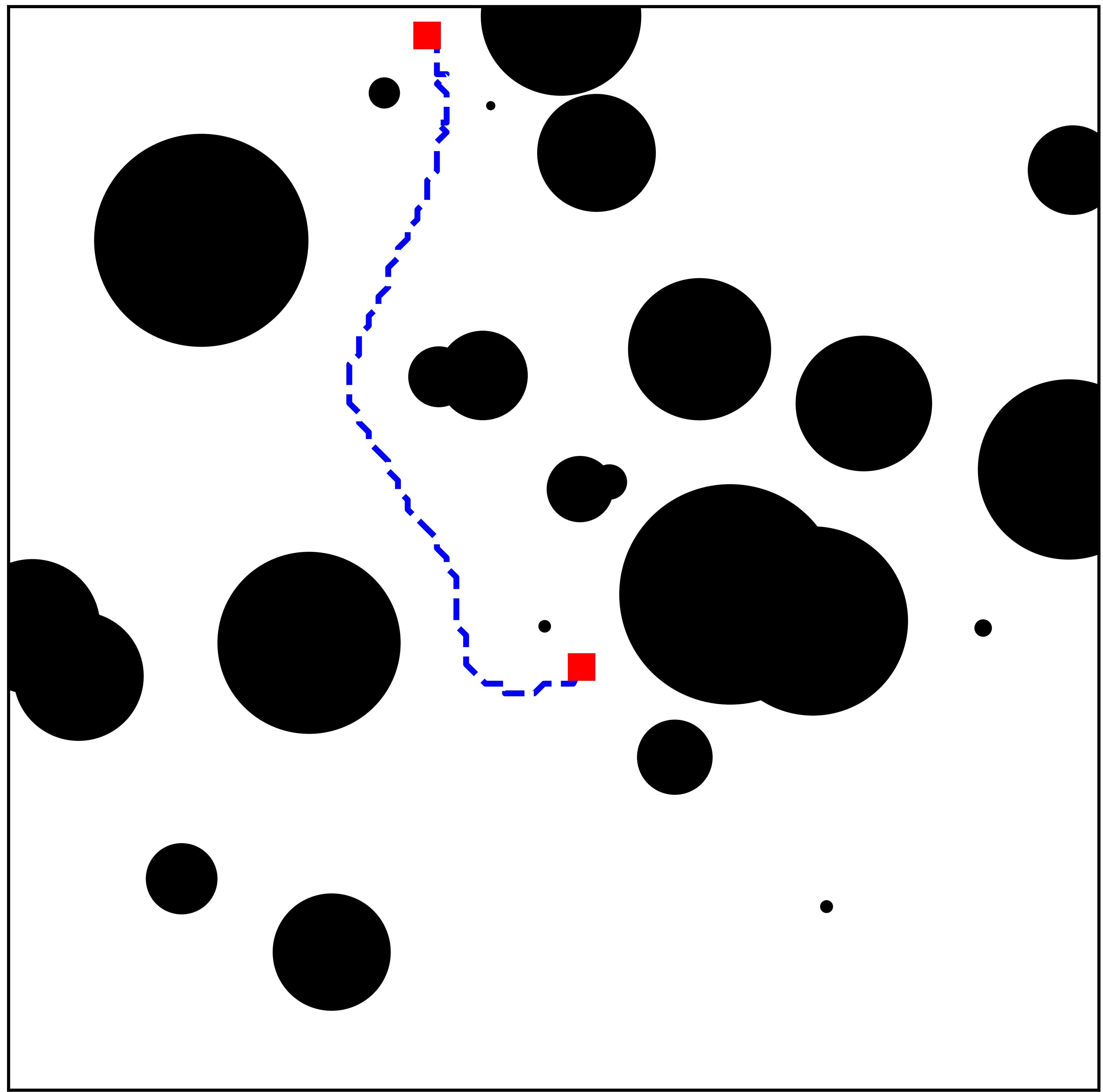}}
    \caption{Solutions found by PPNet under different clearance settings. (a), (b), (c), and (d) clearance=1. (e), (f), (g), and (h) clearance=3. }
    \label{fig_clearance}
\end{figure*}

\subsection{EDaGe-PP Comparison with Popular Data Generation Methods}\label{sec:exp:data}
In this section, we evaluate EDaGe-PP against the popular methods that use different classical planners for finding solutions to random path planning problems, including RRT*, IRRT*, BIT*, and ABIT*, which are represented by PPNet-RRT*, PPNet-IRRT*, PPNet-BIT*, PPNet-ABIT* respectively. Due to the computational inefficiency of classical planners, the size of the dataset used in this section is 10,000.
\par In the comparison of data generation methods, we divide the data generation time into two parts, which are the time for generating solutions and the time for generating data with the format for training of PPNet.  For evaluating the time of generating solutions, we let RRT*, IRRT*, BIT*, and ABIT* run until they find a solution of Euclidean cost within the 5\% range of the cost of the PPNet solution. As shown in Table~\ref{tab:data}, the data generation methods that apply RRT*, IRRT*, BIT*, and ABIT* for finding solutions have about $27 \times $, $10 \times $, $6 \times $, $6 \times $ computation time for generating solutions respectively compared to EDaGe-PP, and have about $33 \times $, $33 \times $, $38 \times $, $38 \times $ computation time for generating data respectively compared to EDaGe-PP. For the cost of the solutions, We report mean computation times with standard deviation over 10,000 path planning problems. Due to the solutions that are generated by EDaGe-PP being continuous-curvature, PPNet can not output solutions comprised of broken lines, which makes the cost of the solutions that are found by PPNet-EDaGe-PP slightly higher than others. However, because of the continuous-curvature solutions, the success rate of PPNet-EDaGe-PP is about $2 \times $ compared to the other methods.

\subsection{Path Planning Under Different Clearances}\label{sec:exp:c}
In this section, we test the performance of our PPNet on the path planning problems under different clearances. 
\par Two kinds of clearance (1, 3) are used. We generate two sets of data with each clearance requirement (1, 3). Then using the two datasets train PPNet respectively. The structure of PPNet this section uses is NAT-UPerHead-64. Fig.\ref{fig_clearance} presents four different path planning problems, which have divergent optimal solutions with different clearance requirements. As shown in Fig.\ref{fig_clearance}, PPNet can find different near-optimal solutions that satisfy different clearance requirements respectively.

\begin{table*}[hb]
    \caption{\textsc{Computation Time, Path Cost, and Success Rate Comparison of PPNet and MPNetPath}}
    \centering 
    \begin{threeparttable}
    \begin{tabular}{|c|c|c|c|c|}\hline
    Methods& Time& Path cost& Success rate$(\%)$\\\hline
        MPNetPath-RRT* &$0.110 \pm 0.123$  &$43.8 \pm 10.8$&$29.75$  \\
        MPNetPath-IRRT* &$0.137 \pm 0.131$  &$52.7 \pm 16.1$&$53.68$ \\
        MPNetPath-BIT* &$0.131 \pm 0.141$  &$48.3 \pm 13.7$&$40.90$\\
        MPNetPath-ABIT* &$0.126 \pm 0.139$  &$47.3 \pm 13.4$&$40.35$\\
        MPNetPath-EDaGe-PP &$0.115 \pm 0.156$  &$42.8 \pm 10.7$&$24.48$ \\\hline
        PPNet &$\mathbf{0.015}$ &$\mathbf{39.3 \pm 7.30}$&$\mathbf{94.68}$  \\\hline
    \end{tabular}
    \begin{tablenotes}
    \footnotesize
    \item[*]The Variants of MPNetPath are trained by datasets that were generated by data generation methods that solutions were found by RRT*, IRRT*, BIT*, and ABIT* respectively.
      \end{tablenotes}
    \end{threeparttable}
    \label{tab:mpnet}
\end{table*}

\subsection{PPNet Comparison With Learning-Based Planner}\label{sec:exp:mpnet}
In this section,  we evaluate PPNet against the representative learning-based planner MPNet. We choose MPNetPath:NP (B), called MPNetPath in this paper, which is the variant with the best performance of all variants of MPNet in computation time, for the experiments in this section. For following the settings of the dataset of MPNet, the size of the dataset used in this section is 4,000, but the dataset contains 4,000 different maps, which is much larger than the setting in the experiments of MPNet (number of maps: 10). The maximum number of iterations of MPNetPath is set to 130, and the part of replanning is 50.
\par Table~\ref{tab:mpnet} present the numerical result of comparison of PPNet and the five variants of MPNetPath that are trained by the datasets generated by EDaGe-PP and the methods that apply classical planners, including RRT*, IRRT*, BIT*, and ABIT*, which are represented by MPNetPath-EDaGe-PP, MPNetPath-RRT*, MPNetPath-IRRT*, MPNetPath-BIT*, MPNetPath-ABIT* respectively. As shown in Table~\ref{tab:mpnet}, PPNet outperforms all variants of MPNetPath in computation time, path cost, and success rate. In all variants of MPNetPath, MPNetPath-PPNet outperforms other variants in path cost. But MPNetPath-EDaGe-PP obtains the worst performance in success rate. The reason is RNN that the neural networks used in MPNet have the limitation in predicting long sequence results, and the solutions generated by EDaGe-PP have more waypoints than other methods. However, this feature is suitable for the methods that apply the neural networks with structures for CV, which is exhibited in the success rate of PPNet which is about $2 \times$ compared to MPNetPath. As for computation time, because of the end-to-end way that PPNet uses for solving path planning problems, PPNet provides about $7 \times$, $9 \times$, $9 \times$, $8 \times$, and $8 \times$ computation speed improvements compared to MPNetPath-RRT*, MPNetPath-IRRT*, MPNetPath-BIT*, MPNetPath-ABIT*, MPNetPath-EDaGe-PP respectively.

\subsection{PPNet Comparison With Sampling-Based Planners}\label{sec:exp:alg}
In this section, we evaluate PPNet against the sampling-based planners including RRT*, IRRT*, BIT*, and ABIT*. PPNet was compared
to the OMPL implementations of RRT*, IRRT*, BIT*, and ABIT*.
\par
\begin{figure*}[t]
    \centering
	\subfloat[\label{fig.alg1.ppnet}]{
		\includegraphics[height=0.18\textwidth]{figs/s_9731_PPNet.jpg}}
    \subfloat[\label{fig.alg1.rrt}]{
		\includegraphics[height=0.18\textwidth]{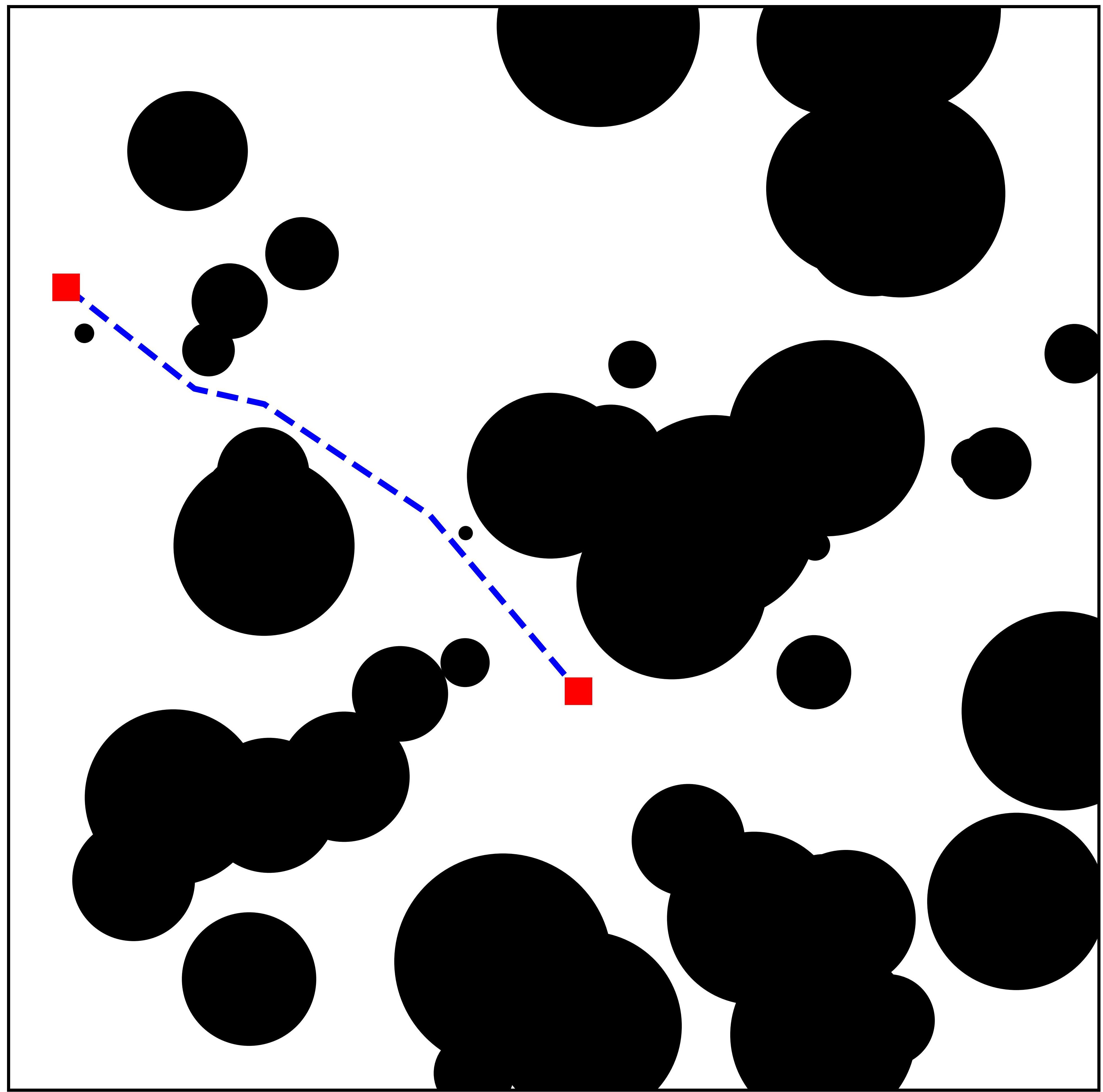}}
    \subfloat[\label{fig.alg1.irrt}]{
		\includegraphics[height=0.18\textwidth]{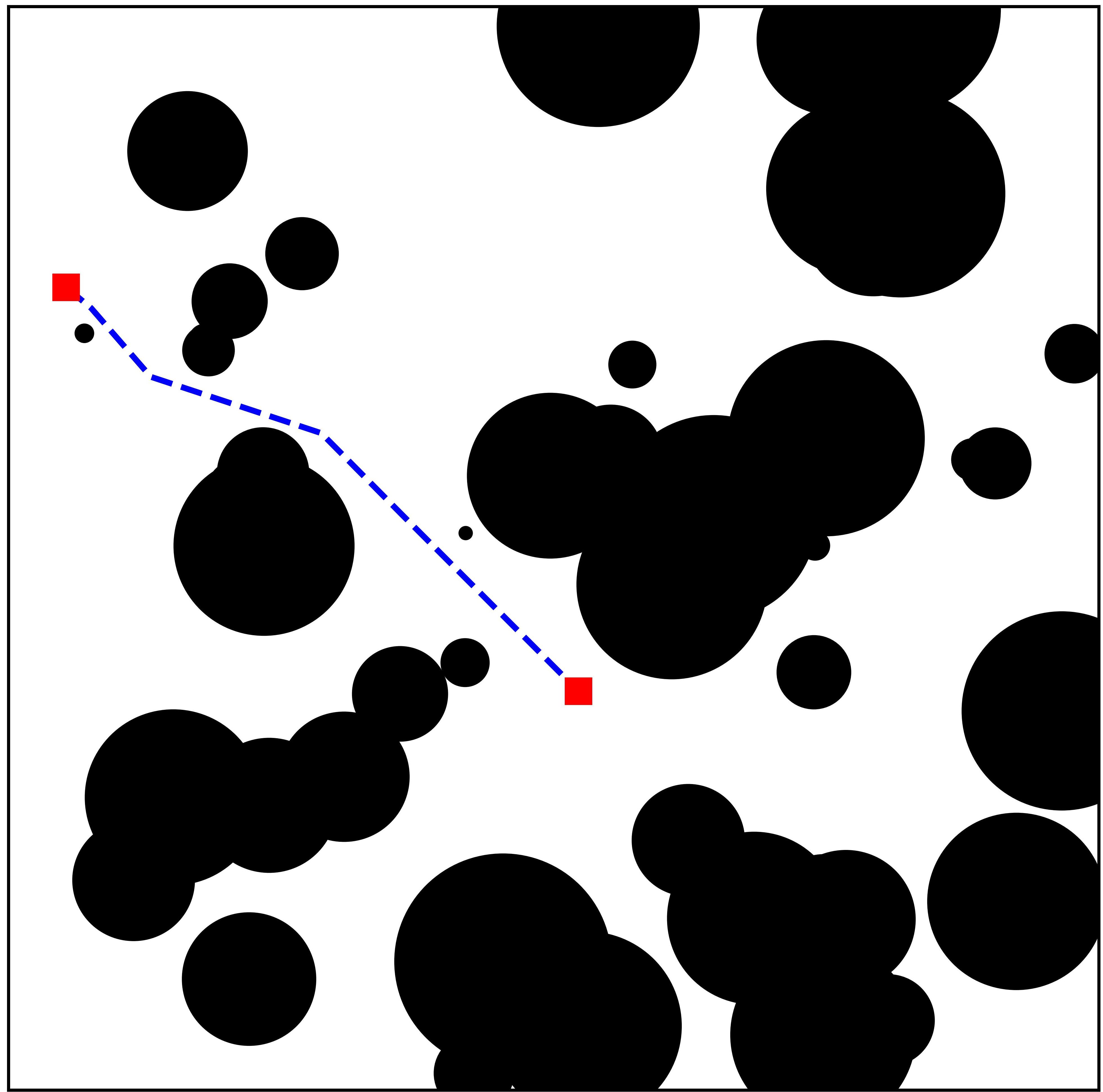}}
    \subfloat[\label{fig.alg1.bit}]{
		\includegraphics[height=0.18\textwidth]{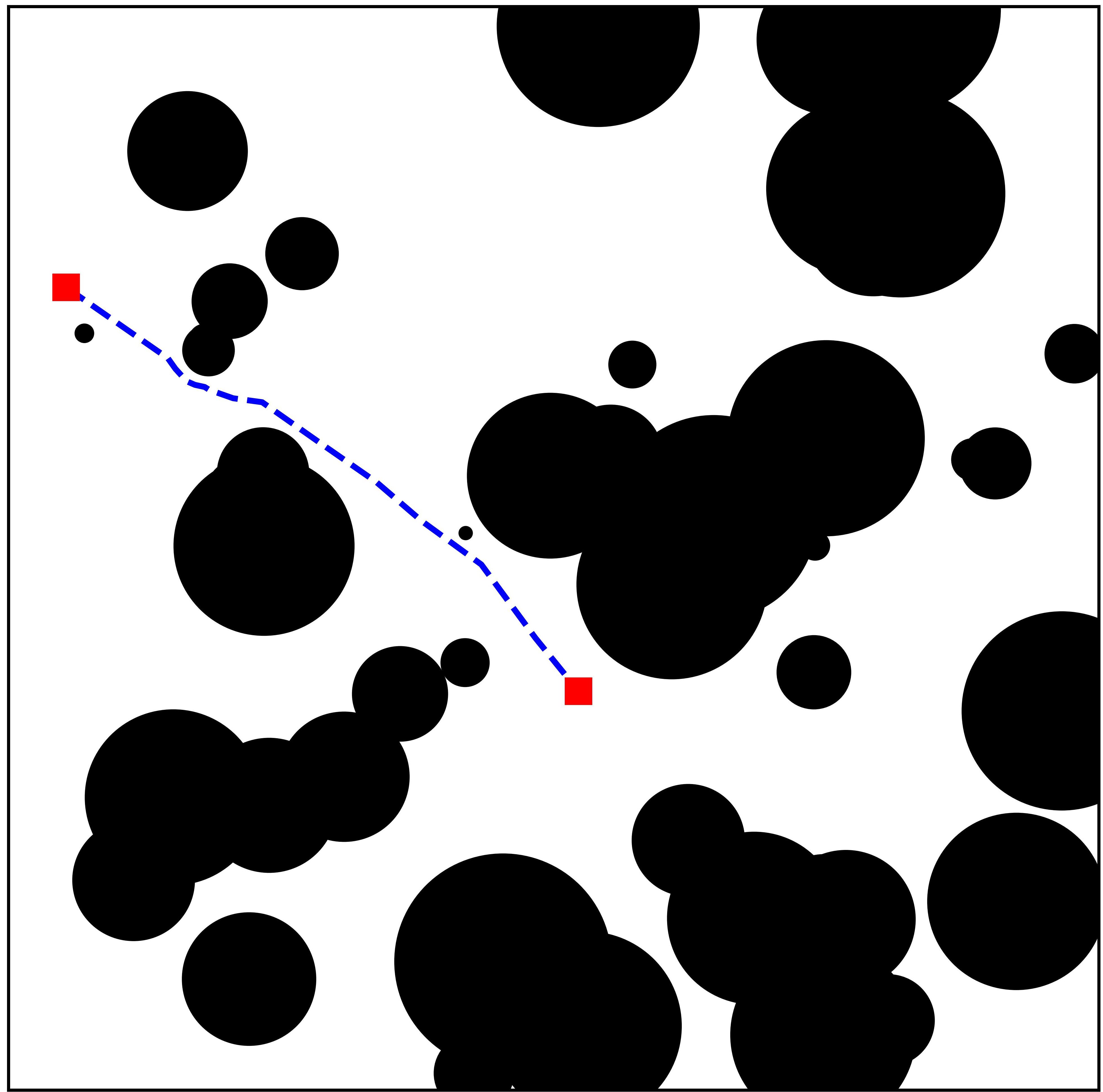}}
    \subfloat[\label{fig.alg1.abit}]{
		\includegraphics[height=0.18\textwidth]{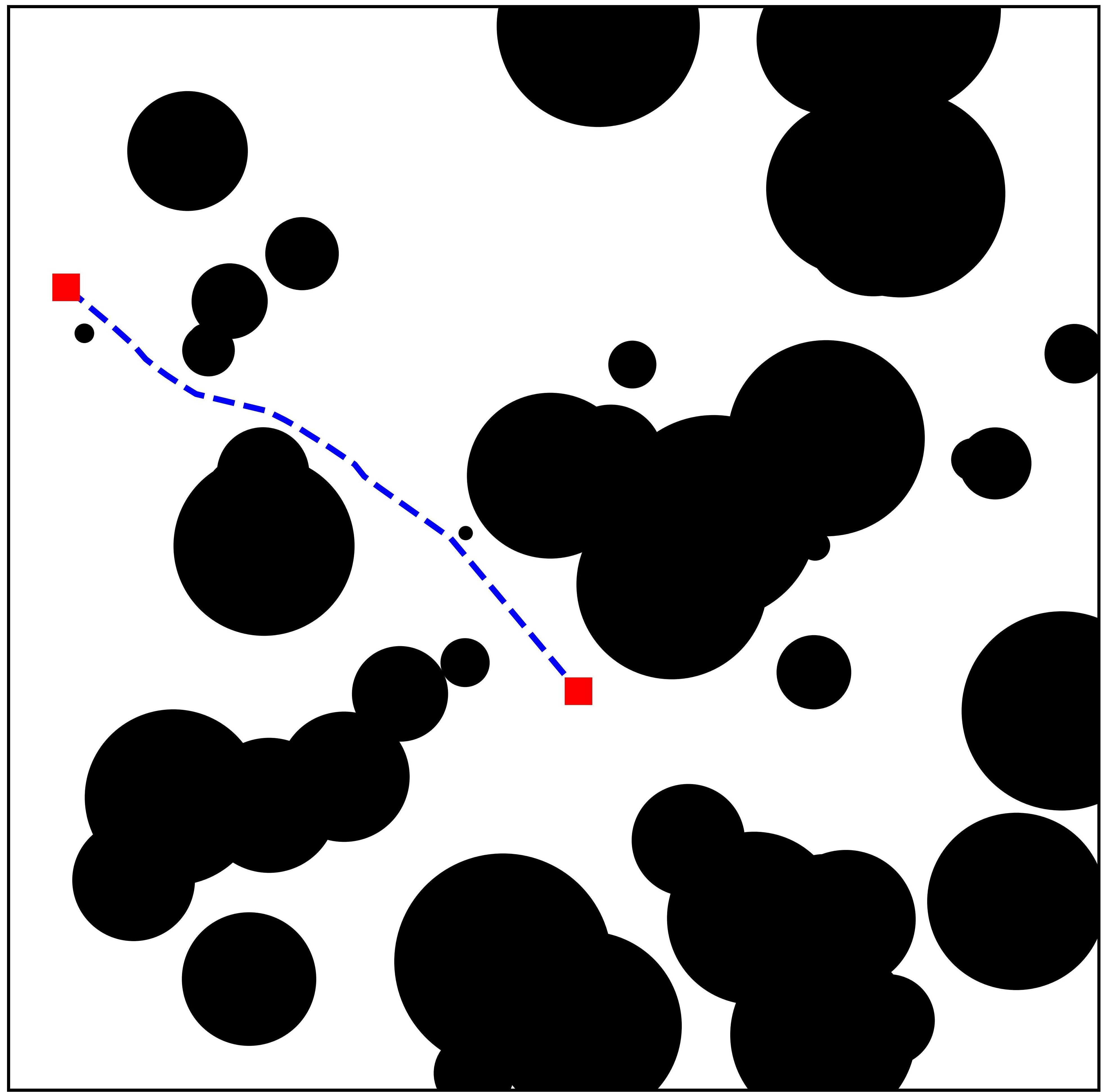}}
    \caption{An example of PPNet, RRT*, IRRT*, BIT*, and ABIT* run on a random $ \mathbb{R}^{2}$ world. Each algorithm was run until it found an equivalent solution to PPNet ($c=31.49$). (a) $t=0.015\mathrm{s}$. (b) $t=2.452\mathrm{s}$. (c) $t=0.580\mathrm{s}$. (d) $t=60.001\mathrm{s}$. (e) $t=9.564\mathrm{s}$. }
    \label{fig.alg1}
\end{figure*}

 We let RRT*, IRRT*, BIT*, and ABIT* run until they find an equivalent solution to PPNet and a solution of Euclidean cost within the 2\%, and 5\% range of the cost of the PPNet solution respectively. We report mean computation times with a standard deviation over 10,000 random path planning problems. PPNet can find a solution within a specific planning time, which is 15.3ms. The state-of-the-art classical planners, RRT*, IRRT*, BIT*, and ABIT* exhibit shorter computation times in simple planning problems, but higher mean computation times than PPNet. Furthermore, standard deviations of computation times are high, which is unacceptable in some tasks (Fig.\ref{fig.alg1}). As shown in Table~\ref{tab:algo}, RRT*, IRRT*, BIT*, and ABIT* need about $15 \times$, $8 \times$, $8 \times$, and $8 \times$ computation time for finding an equivalent solution to PPNet respectively. And Fig.\ref{fig.alg1} presents a difficult case for sampling-based planners. For solving this path planning problem, RRT*, IRRT*, BIT*, and ABIT* need $2.452\mathrm{s}$, $0.580\mathrm{s}$, $60.001\mathrm{s}$, and $9.564\mathrm{s}$  of computation time. And PPNet can still output the near-optimal solution in 15.3ms.

\section{Conclusion}\label{sec:clusion}
In this article, we also propose PPNet, which is a two-stage neural network in which each stage solves one of the two subproblems of the path planning problem. Moreover, we also propose EDaGe-PP, an efficient data generation method for path planning. Due to the continuous-curvature path with analytical expression, the dataset is beneficial for the training of neural networks. PPNet can find a near-optimal solution in an end-to-end way within 15.3ms.
\par In future work, we will try to improve our neural network structure to develop an end-to-end near-optimal path planner in 3D space. In addition, extending EDaGe-PP to 3D space is also challenging work. Code is available at: \href{https://github.com/AdamQLMeng/PPNet}{https://github.com/AdamQLMeng/PPNet}.

\begin{table*}[hb]
    \caption{\textsc{Computation Time Comparison of PPNet and RRT*, IRRT*, BIT*, ABIT*}}
    \centering
    \begin{threeparttable}
    \begin{tabular}{|c|c|c|c|}
        \hline
        Methods& $+0\%$ PPNet cost& $+2\%$ PPNet cost& $+5\%$ PPNet cost\\\hline
        RRT*  &  $0.219 \pm 0.278$   &  $0.159 \pm 0.226$  &  $0.119 \pm 0.181$   \\
        IRRT*  &  $0.116 \pm 0.152$  &  $0.099 \pm 0.141$  &  $0.082 \pm 0.116$    \\
        BIT* &  $0.118 \pm 1.088$    &  $0.046 \pm 0.113$  &  $0.031 \pm 0.035$   \\
        ABIT* &  $0.116 \pm 1.042$   &  $0.045 \pm 0.063$  &  $0.031 \pm 0.060$    \\\hline
        PPNet & \multicolumn{3}{c|}{$\mathbf{0.015}$}\\
        \hline
    \end{tabular}
    \begin{tablenotes}
    \footnotesize
    \item[*]RRT*, IRRT*, BIT*, and ABIT* were run until they found an equivalent solution to PPNet and a solution of Euclidean cost within the 2\%, 5\% range of the cost of the PPNet solution.
    \end{tablenotes}
    \end{threeparttable}
    \label{tab:algo}
\end{table*}

\begin{figure*}[ht]
    \centering
	\subfloat[\label{fig.alg2.ppnet}]{
		\includegraphics[height=0.18\textwidth]{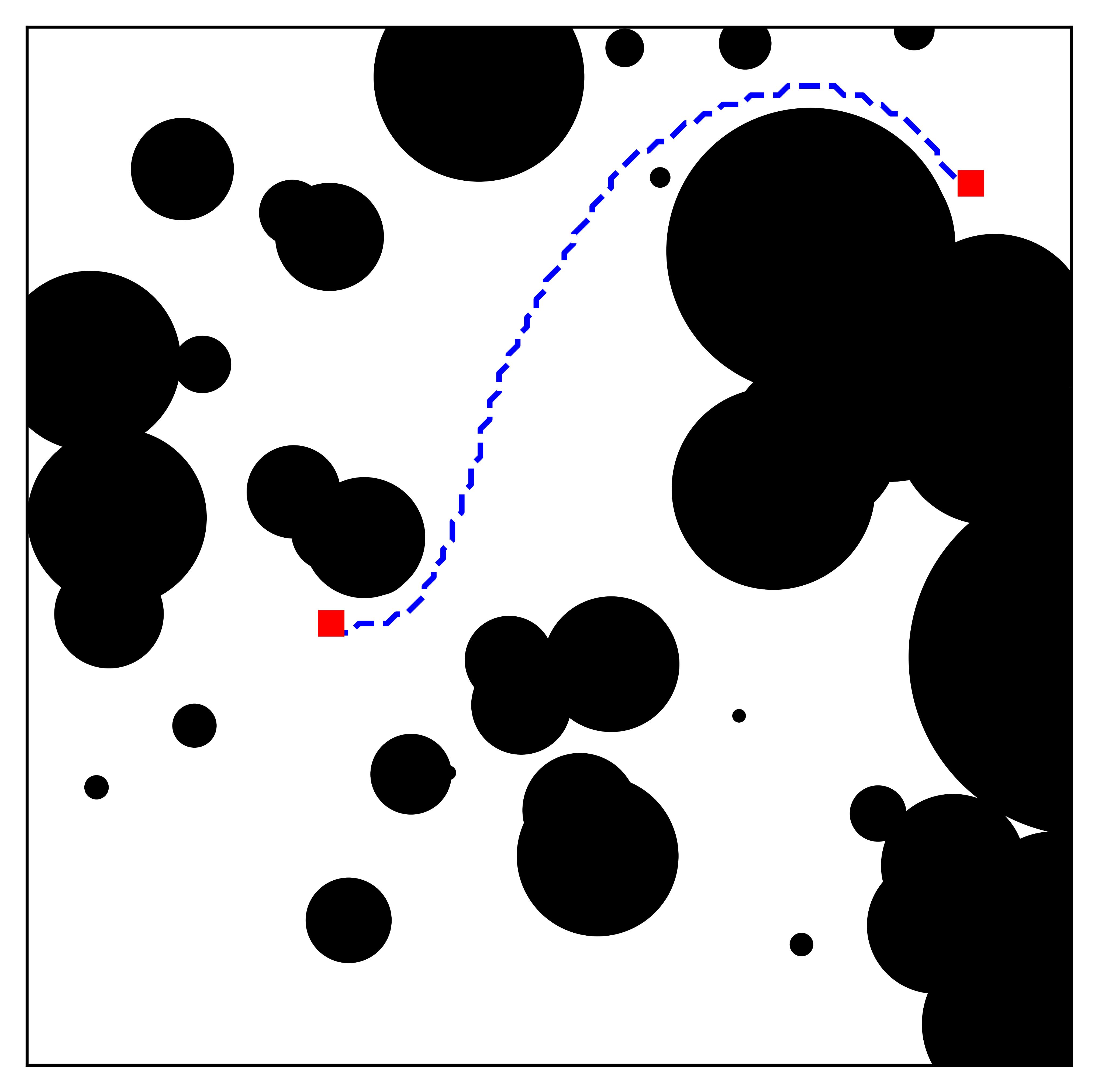}}
    \subfloat[\label{fig.alg2.rrt}]{
		\includegraphics[height=0.18\textwidth]{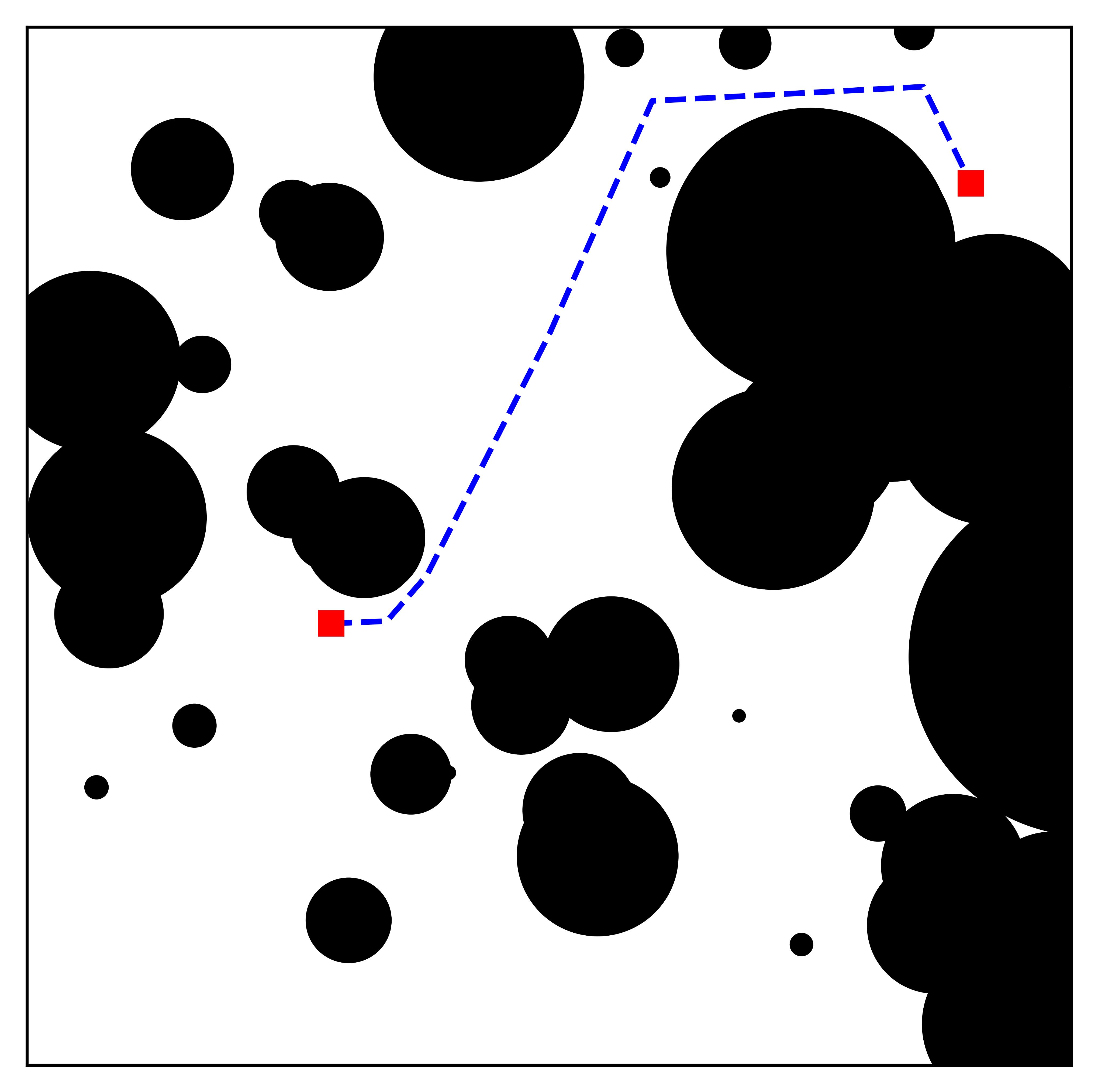}}
    \subfloat[\label{fig.alg2.irrt}]{
		\includegraphics[height=0.18\textwidth]{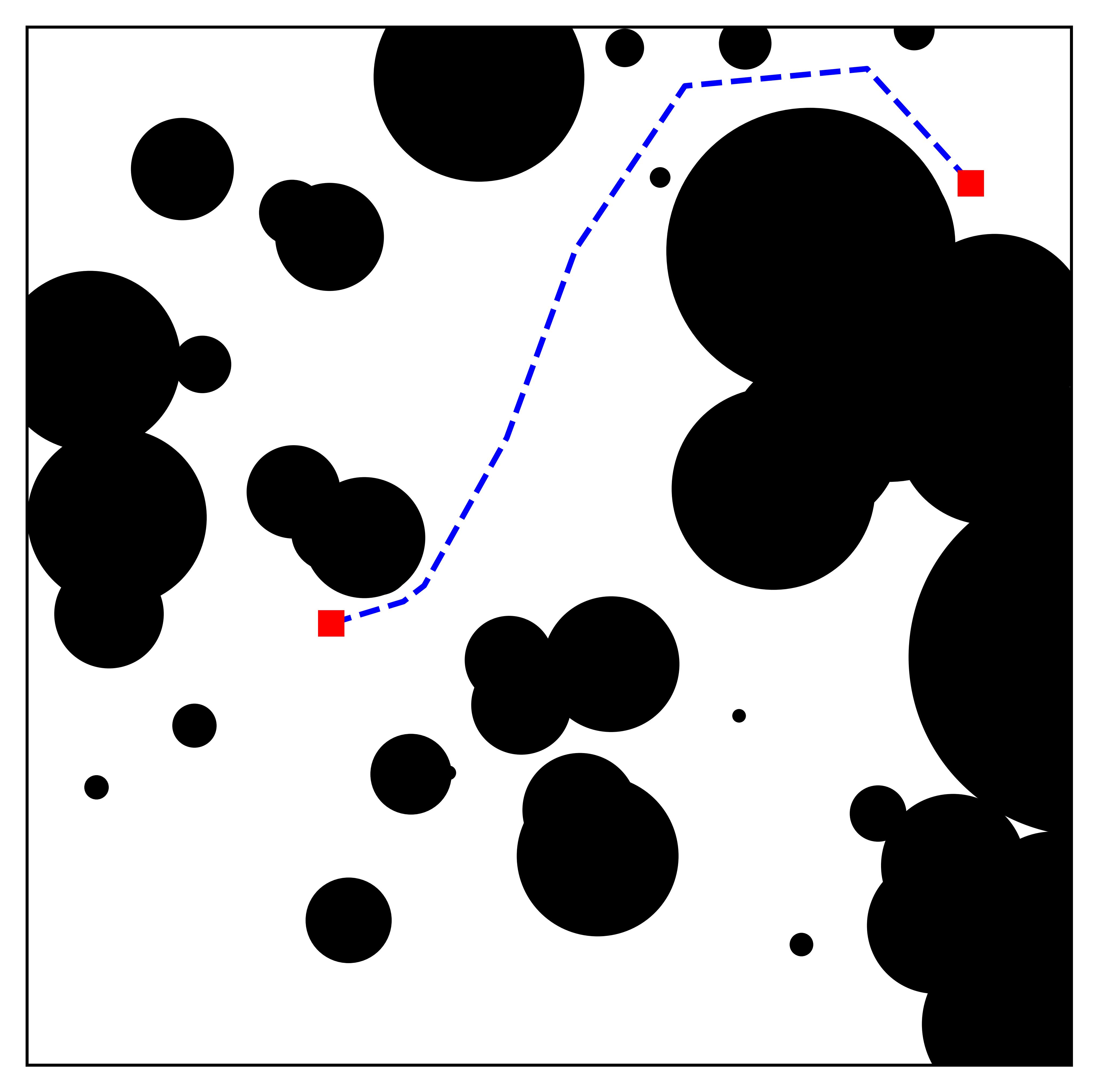}}
    \subfloat[\label{fig.alg2.bit}]{
		\includegraphics[height=0.18\textwidth]{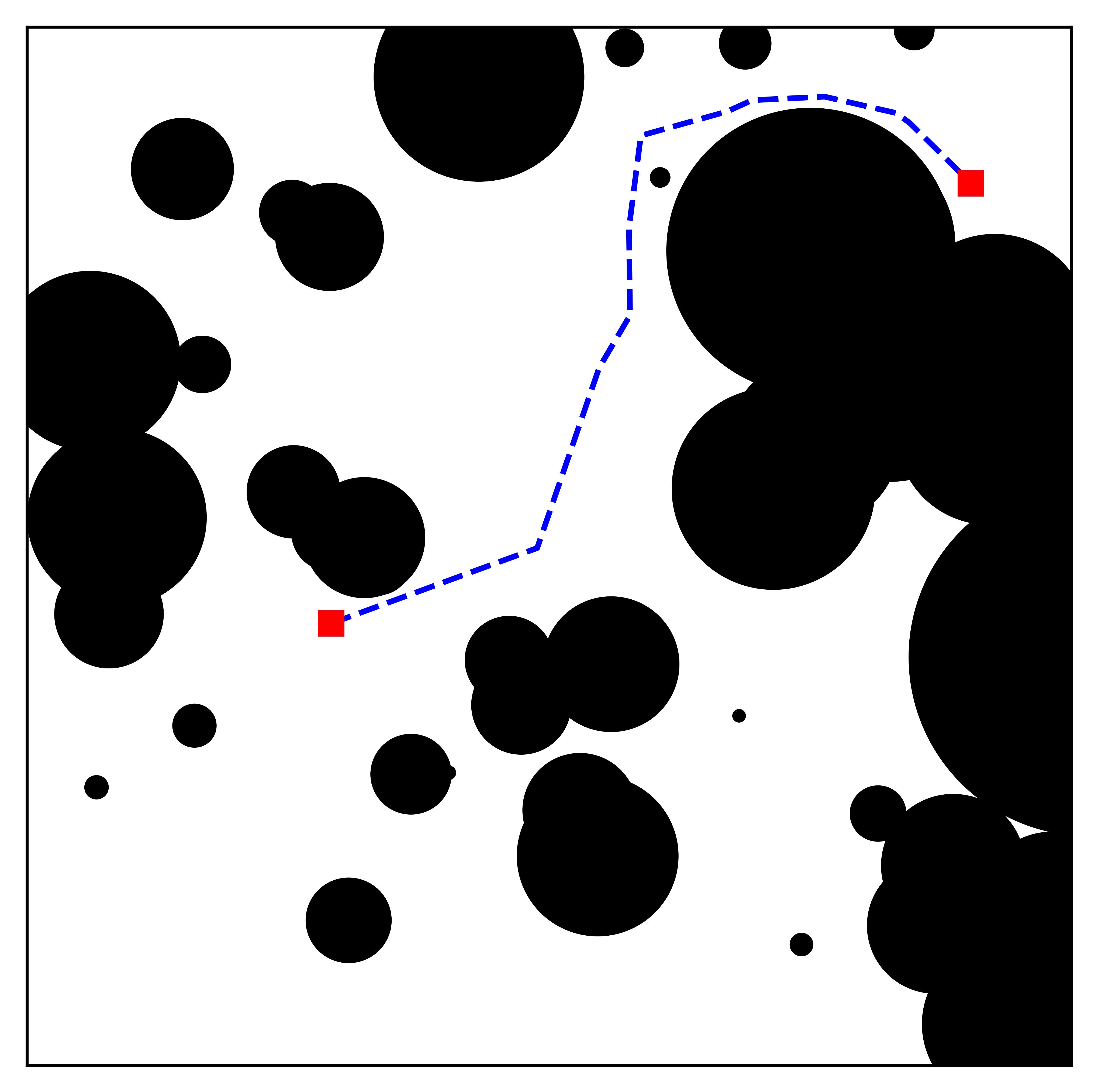}}
    \subfloat[\label{fig.alg2.abit}]{
		\includegraphics[height=0.18\textwidth]{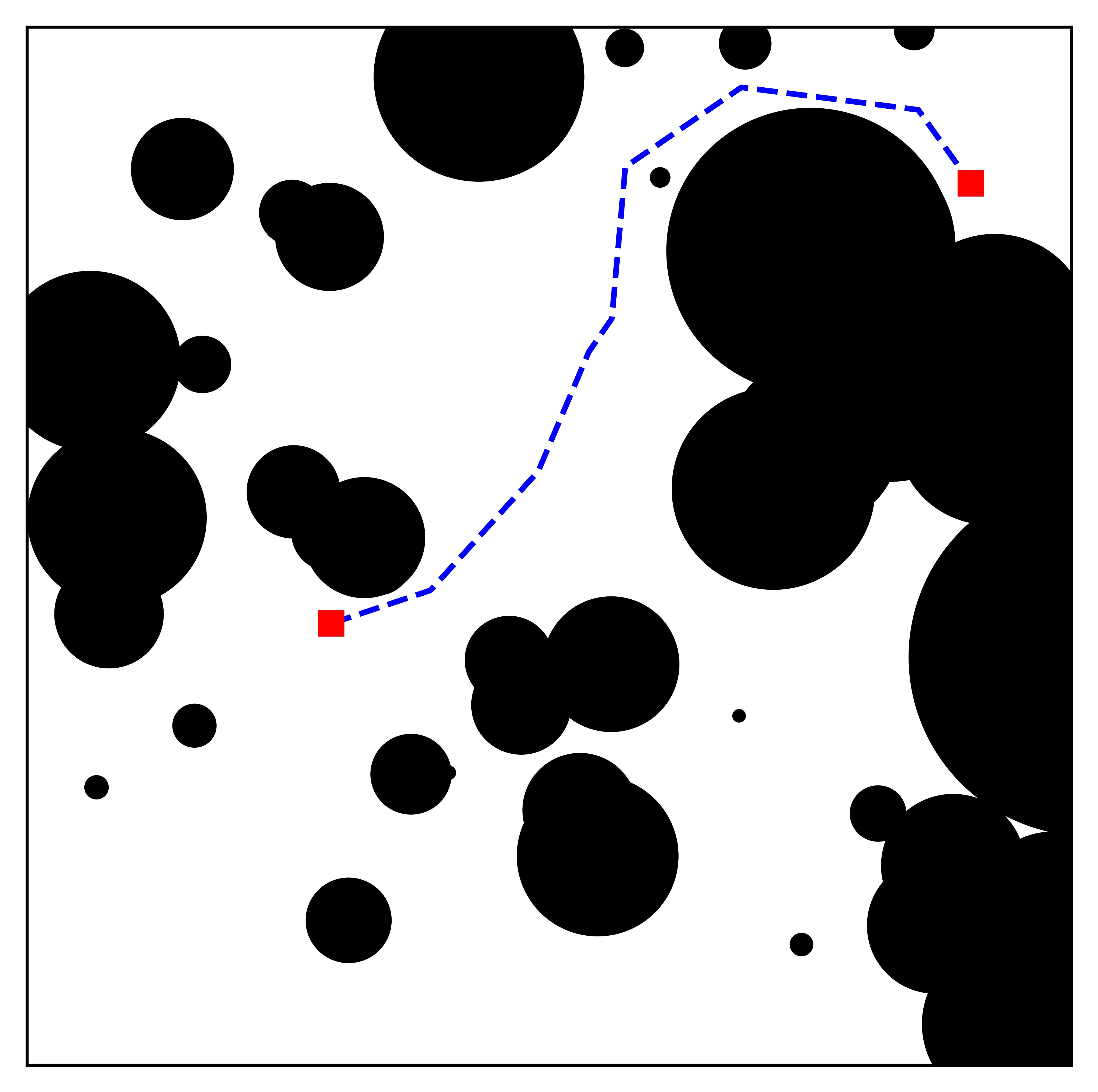}}
    \caption{An example of PPNet, RRT*, IRRT*, BIT*, and ABIT* run on a u-turn environment. Each algorithm was run until it found an equivalent solution to PPNet ($c=50.03$). (a) $t=0.015\mathrm{s}$. (b) $t=0.539\mathrm{s}$. (c) $t=0.503\mathrm{s}$. (d) $t=0.220\mathrm{s}$. (e) $t=0.104\mathrm{s}$. }
    \label{fig.alg2}
\end{figure*}

\begin{IEEEbiography}[{\includegraphics[width=1in,height=1.25in,clip,keepaspectratio]{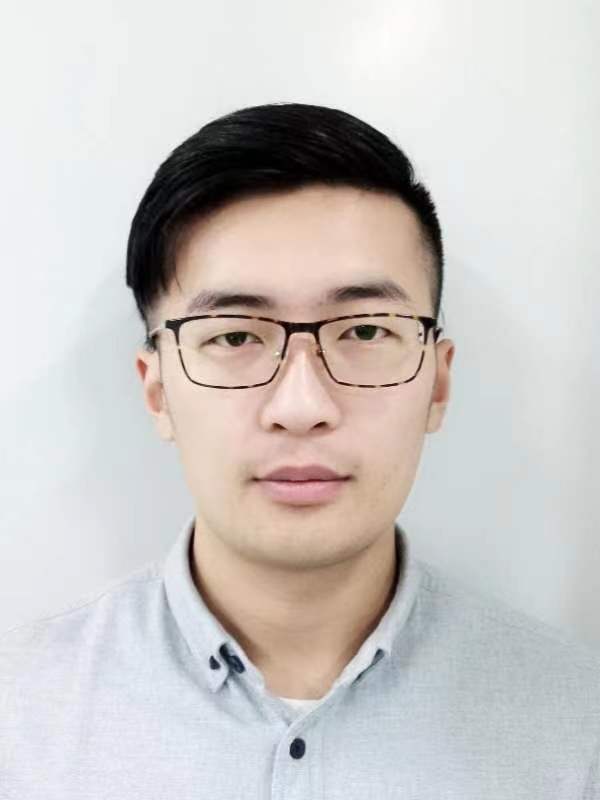}}]{Qinglong Meng}
received the B.E. degree in measurement and control technology and instrumentation from the College of Instrumentation and Electrical Engineering, Jilin University, Changchun, China, in 2017, He is currently working toward the M.S. degree in electronic information with Tsinghua Shenzhen International Graduate School, Tsinghua University, Shenzhen, China.
\par His research interests include robotics, motion planning, and machine
learning in robotics.
\end{IEEEbiography}

\begin{IEEEbiography}[{\includegraphics[width=1in,height=1.25in,clip,keepaspectratio]{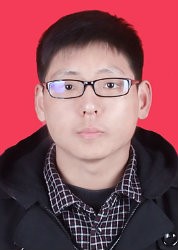}}]{Chongkun Xia}
received the Ph.D. degree in pattern recognition and intelligent systems from Northeastern University, Shenyang, China, in 2021.
\par He is a Postdoctor with the Centre for Artificial Intelligence and Robotics, Tsinghua Shenzhen International Graduate School, Tsinghua University, Shenzhen, China. His research interests include robotics, motion planning, and machine learning.
\end{IEEEbiography}

\begin{IEEEbiography}[{\includegraphics[width=1in,height=1.25in,clip,keepaspectratio]{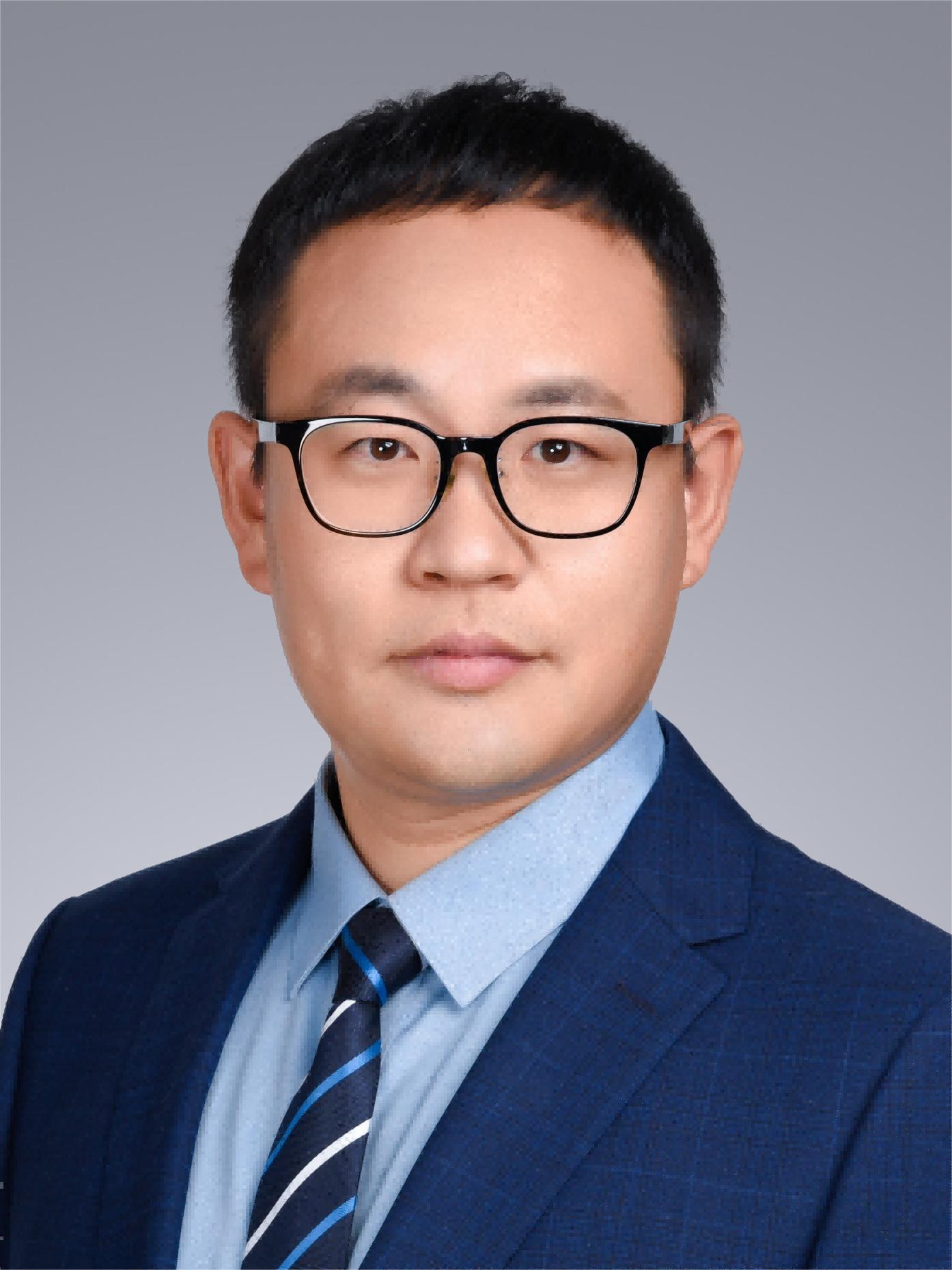}}]{Xueqian Wang}
received the B.E.
degree in mechanical design, manufacturing, and automation from the Harbin University of Science and
Technology, Harbin, China, in 2003, and the M.Sc.
degree in mechatronic engineering and the Ph.D.
degree in control science and engineering from the
Harbin Institute of Technology (HIT), Harbin, China,
in 2005 and 2010, respectively.
\par From June 2010 to February 2014, he was the Post-
doctoral Research Fellow with HIT. He is currently a
Professor and the Leader of the Center of Intelligent
Control and Telescience, Tsinghua Shenzhen International Graduate School, Tsinghua University, Shenzhen, China. His research interests include dynamics modeling, control, and teleoperation of robotic systems.
\end{IEEEbiography}

\begin{IEEEbiography}[{\includegraphics[width=1in,height=1.25in,clip,keepaspectratio]{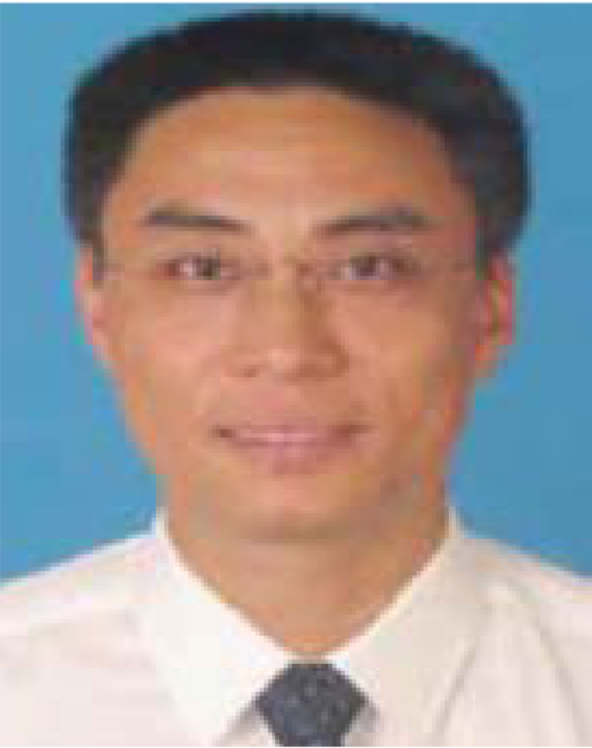}}]{Bin Liang}
received the Ph.D. degree in precision instrument and mechanology from Tsinghua University, Beijing, China, in 1994.,He is currently a Professor with the Department of Automation, Tsinghua University. His research interests include space robotics, manipulators, and intelligent control.
\end{IEEEbiography}
\vspace{-400pt}
\begin{IEEEbiography}[{\includegraphics[width=1in,height=1.25in,clip,keepaspectratio]{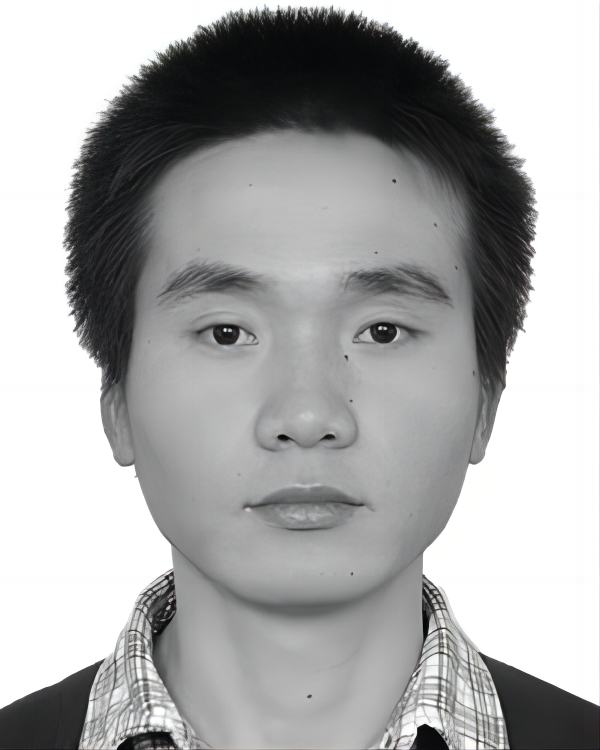}}]{Songping Mai}
received the B.S. degree in electronic information from Wuhan University, Hubei, China, in 2003, and the Ph.D. degree in electronic engineering from Tsinghua University, Beijing, China, in 2008.,Since 2008, he has been with the Integrated Circuits and Systems Design Laboratory, Graduate School at Shenzhen, Tsinghua University, Shenzhen, China, where he is currently an Associate Professor. His research interests include robotics, artificial intelligence, and circuit design.
\end{IEEEbiography}

\begin{thebibliography}{1}
\bibitem{rrt*}
 S. Karaman and E. Frazzoli, “Sampling-based algorithms for optimal motion planning,” The International Journal of Robotics Research, vol. 30, no. 7, pp. 846–894, 2011.
  \bibitem{irrt*}
J. D. Gammell, S. S. Srinivasa and T. D. Barfoot, "Informed RRT*: Optimal sampling-based path planning focused via direct sampling of an admissible ellipsoidal heuristic," 2014 IEEE/RSJ International Conference on Intelligent Robots and Systems, Chicago, IL, USA, 2014, pp. 2997-3004, doi: 10.1109/IROS.2014.6942976.
\bibitem{bit*}
J. D. Gammell, S. S. Srinivasa, and T. D. Barfoot, “Batch Informed Trees (BIT*): Sampling-based optimal planning via the heuristically guided search of implicit random geometric graphs,” in Proceedings of the International Conference on Robotics and Automation (ICRA), 2015, pp. 3067–3074.
\bibitem{abit*}
M. P. Strub and J. D. Gammell, "Advanced BIT* (ABIT*): Sampling-Based Planning with Advanced Graph-Search Techniques," 2020 IEEE International Conference on Robotics and Automation (ICRA), Paris, France, 2020, pp. 130-136, doi: 10.1109/ICRA40945.2020.9196580.
\bibitem{nrrt*}
J. Wang, W. Chi, C. Li, C. Wang and M. Q. . -H. Meng, "Neural RRT*: Learning-Based Optimal Path Planning," in IEEE Transactions on Automation Science and Engineering, vol. 17, no. 4, pp. 1748-1758, Oct. 2020, doi: 10.1109/TASE.2020.2976560.
\bibitem{mpnet}
A. H. Qureshi, A. Simeonov, M. J. Bency and M. C. Yip, "Motion Planning Networks," 2019 International Conference on Robotics and Automation (ICRA), Montreal, QC, Canada, 2019, pp. 2118-2124, doi: 10.1109/ICRA.2019.8793889.


\bibitem{fmt*}
L. Janson, E. Schmerling, A. Clark, and M. Pavone, “Fast Marching Tree: A fast marching sampling-based method for optimal motion planning in many dimensions,” The International Journal of Robotics Research, vol. 34, no. 7, pp. 883–921, 2015.
\bibitem{lpa*}
S. Koenig, M. Likhachev, and D. Furcy, “Lifelong Planning A*,” Artificial Intelligence, vol. 155, no. 1-2, pp. 93–146, 2004.
L. E. Kavraki, P. ˇSvestka, J.-C. Latombe, and M. H.

\bibitem{need}
J. Wang, J. Liu, W. Chen, W. Chi and M. Q. . -H. Meng, "Robot Path Planning via Neural-Network-Driven Prediction," in IEEE Transactions on Artificial Intelligence, vol. 3, no. 3, pp. 451-460, June 2022, doi: 10.1109/TAI.2021.3119890.
\bibitem{3dnrrt*}
J. Wang, X. Jia, T. Zhang, N. Ma and M. Q. . -H. Meng, "Deep Neural Network Enhanced Sampling-Based Path Planning in 3D Space," in IEEE Transactions on Automation Science and Engineering, vol. 19, no. 4, pp. 3434-3443, Oct. 2022, doi: 10.1109/TASE.2021.3121408.
\bibitem{rrt-c}
J. Kuffner and S. LaValle, “RRT-Connect: An efficient approach to single-query path planning,” in Proceedings of the International Conference on Robotics and Automation (ICRA), vol. 2, 2000, pp. 995–1001.
\bibitem{mmseg}
MMSegmentation Contributors. MMSegmentation: Open-mmlab semantic segmentation toolbox and benchmark. https://github.com/open-mmlab/mmsegmentation, 2020.
\bibitem{vit}
Alexey Dosovitskiy, Lucas Beyer, Alexander Kolesnikov, Dirk Weissenborn, Xiaohua Zhai, Thomas Unterthiner, Mostafa Dehghani, et al. An image is worth 16x16 words: Transformers for image recognition at scale. In International Conference on Learning Representations (ICLR), 2020.
\bibitem{swin}
Z. Liu et al., "Swin Transformer: Hierarchical Vision Transformer using Shifted Windows," 2021 IEEE/CVF International Conference on Computer Vision (ICCV), Montreal, QC, Canada, 2021, pp. 9992-10002, doi: 10.1109/ICCV48922.2021.00986.
\bibitem{nat}
Ali Hassani, Steven Walton, Jiachen Li, Shen Li, and Humphrey Shi. Neighborhood attention transformer. arXiv:2204.07143, 2022.
\bibitem{dinat}
Ali Hassani, and Humphrey Shi. Dilated Neighborhood attention transformer. arXiv:2209.15001, 2022.
\bibitem{setr}
S. Zheng et al., "Rethinking Semantic Segmentation from a Sequence-to-Sequence Perspective with Transformers," 2021 IEEE/CVF Conference on Computer Vision and Pattern Recognition (CVPR), Nashville, TN, USA, 2021, pp. 6877-6886, doi: 10.1109/CVPR46437.2021.00681.
\bibitem{sa}
Ashish Vaswani, Noam Shazeer, Niki Parmar, Jakob Uszkoreit, Llion Jones, Aidan N Gomez, Łukasz Kaiser, and Illia Polosukhin. Attention is all you need. In Advances in Neural Information Processing Systems (NeurIPS), 2017.
\bibitem{upernet}
Tete Xiao, Yingcheng Liu, Bolei Zhou, Yuning Jiang, and Jian Sun. "Unified perceptual parsing for scene understanding," in Proceedings of the European Conference on Computer Vision (ECCV), pages 418–434, 2018.
\end{thebibliography}
\end{document}